\pdfoutput=1

\documentclass[11pt]{article}

\PassOptionsToPackage{hyperfootnotes=false}{hyperref}

\usepackage[final]{emnlp2021}

\usepackage{times}
\usepackage{latexsym}
\usepackage{amsmath}
\usepackage{amssymb}
\usepackage{graphicx}
\usepackage{afterpage}
\usepackage{booktabs}
\usepackage[T1]{fontenc}

\usepackage[utf8]{inputenc}

\usepackage{microtype}

\usepackage{xcolor}
\usepackage[most]{tcolorbox}
\usepackage{listings}

\definecolor{promptblue}{HTML}{2C7FB8}
\definecolor{promptbluebg}{HTML}{EAF3FB}
\definecolor{promptgreen}{HTML}{2E7D32}
\definecolor{promptgreenbg}{HTML}{E8F5E9}
\definecolor{promptorange}{HTML}{E65100}
\definecolor{promptorangebg}{HTML}{FFF3E0}
\definecolor{promptred}{HTML}{C62828}
\definecolor{promptredbg}{HTML}{FFEBEE}
\definecolor{promptyellow}{HTML}{F57F17}
\definecolor{promptyellowbg}{HTML}{FFFDE7}
\definecolor{promptgray}{HTML}{37474F}
\definecolor{promptgraybg}{HTML}{ECEFF1}

\lstdefinestyle{promptinner}{
  basicstyle=\scriptsize\ttfamily,
  breaklines=true,
  breakatwhitespace=false,
  columns=flexible,
  keepspaces=true,
  showstringspaces=false,
  upquote=true,
  frame=none,
  aboveskip=0pt,
  belowskip=0pt,
  xleftmargin=0pt,
  xrightmargin=0pt,
  extendedchars=true,
  literate=%
    {—}{{\textemdash}}1
    {–}{{\textendash}}1
    {→}{{$\rightarrow$}}1
    {×}{{$\times$}}1
    {≈}{{$\approx$}}1
}

\tcbset{promptstyle/.style={
  breakable, enhanced,
  width=\linewidth,
  boxsep=3pt, left=1mm, right=1mm, top=1mm, bottom=1mm,
  fonttitle=\bfseries\small\color{white},
  attach boxed title to top left={yshift=-2.5mm, xshift=2mm},
  boxed title style={sharp corners, left=1mm, right=1mm},
}}

\newtcblisting{pblue}[1]{promptstyle,
  colback=promptbluebg, colframe=promptblue,
  colbacktitle=promptblue,
  boxed title style={colback=promptblue, colframe=promptblue,
                     sharp corners, left=1mm, right=1mm},
  title={\textsf{#1}},
  listing only,
  listing options={style=promptinner}}

\newtcblisting{pgreen}[1]{promptstyle,
  colback=promptgreenbg, colframe=promptgreen,
  colbacktitle=promptgreen,
  boxed title style={colback=promptgreen, colframe=promptgreen,
                     sharp corners, left=1mm, right=1mm},
  title={\textsf{#1}},
  listing only,
  listing options={style=promptinner}}

\newtcblisting{porange}[1]{promptstyle,
  colback=promptorangebg, colframe=promptorange,
  colbacktitle=promptorange,
  boxed title style={colback=promptorange, colframe=promptorange,
                     sharp corners, left=1mm, right=1mm},
  title={\textsf{#1}},
  listing only,
  listing options={style=promptinner}}

\newtcblisting{pred}[1]{promptstyle,
  colback=promptredbg, colframe=promptred,
  colbacktitle=promptred,
  boxed title style={colback=promptred, colframe=promptred,
                     sharp corners, left=1mm, right=1mm},
  title={\textsf{#1}},
  listing only,
  listing options={style=promptinner}}

\newtcblisting{pyellow}[1]{promptstyle,
  colback=promptyellowbg, colframe=promptyellow,
  colbacktitle=promptyellow,
  boxed title style={colback=promptyellow, colframe=promptyellow,
                     sharp corners, left=1mm, right=1mm},
  title={\textsf{#1}},
  listing only,
  listing options={style=promptinner}}

\newtcblisting{pgray}[1]{promptstyle,
  colback=promptgraybg, colframe=promptgray,
  colbacktitle=promptgray,
  boxed title style={colback=promptgray, colframe=promptgray,
                     sharp corners, left=1mm, right=1mm},
  title={\textsf{#1}},
  listing only,
  listing options={style=promptinner}}

\title{PhyDrawGen: Physically Grounded Diagram Generation from Natural Language}

\makeatletter
\newcommand{\nosymbolthanks}[1]{%
  \protected@xdef\@thanks{\@thanks\protect\footnotetext[0]{#1}}}
\makeatother

\author{
  Nafiul Haque\thanks{\ Equal contribution.}\quad
  Syed Nazmus Sakib\footnotemark[1]\quad
  Shifat E Arman\thanks{\ Corresponding author: \texttt{shifatearman@du.ac.bd}}%
  \nosymbolthanks{Under review at EMNLP. This is a preprint of the submitted manuscript.} \\
  Department of Robotics and Mechatronics Engineering, University of Dhaka \\}

\begin{document}
\maketitle
\begin{abstract}
Generating physics diagrams from text requires strict adherence to physical laws. While current generative models produce visually plausible outputs, they systematically hallucinate force vectors, ignore conservation laws, and violate geometric constraints. We present \textbf{PhyDrawGen}, a neuro-symbolic pipeline that decouples semantic scene understanding from physical constraint satisfaction. First, a large language model extracts a typed scene graph from the problem text. A deterministic solver then converts this graph into a Planar Straight-Line Graph~(PSLG), encoding force balance, optical paths, and field topologies as exact geometric primitives. Finally, a fine-tuned Qwen-VL model implements a visually grounded propose-verify loop to iteratively correct any constraint violations. Evaluated on a benchmark of 1,449 problems spanning mechanics, optics, and electromagnetism, \textbf{PhyDrawGen} significantly outperforms GPT-5-image, Gemini~2.5~Flash, and Gemini~3~Pro, demonstrating robust physical accuracy even on unusual-object problems.
\end{abstract}
\section{Introduction}
Physics diagrams, free-body diagrams, ray-optics
constructions, and electromagnetic field maps are formal visual arguments in which every arrow encodes a physical law, every angle encodes a geometric constraint, and every spatial relationship encodes an interaction governed by classical physics. A force arrow pointing in the wrong direction is not merely an aesthetic defect; it is a false physical claim. As such the generation of physically grounded scientific diagrams represents a frontier at the intersection of natural language understanding, structured reasoning, and constrained visual synthesis.

Though diffusion models are capable of producing photorealistic images~\citep{rombach2022ldm,ramesh2022dalle2,saharia2022photorealistictexttoimagediffusionmodels,ho2020ddpm} and spatially conditioned generation through adapters~\citep{zhang2023controlnet,mou2023t2iadapter,ye2023ipadapter} and grounding mechanisms~\citep{li2023gligen,bar2023multidiffusion,johnson2018sg2im}, the noise-addition and denoising architecture of diffusion models is fundamentally ill-suited to tasks that require hard constraint satisfaction. The denoising process optimizes for perceptual plausibility under a learned prior, not for algebraic correctness under physical law. When applied to physics diagram generation, this manifests as systematic failure: diffusion models hallucinate force directions, place arrows at geometrically inconsistent angles, omit forces that conservation laws require, and conflate visually similar but physically distinct configurations such as static friction opposing motion versus kinetic friction during sliding.

Recent benchmarks have documented the remarkable ability of large language models (LLMs) and vision-language models (VLMs) to \emph{solve} physics problems from diagrams~\citep{he2024olympiadbench,xiang2025seephysdoesseeinghelp,lu2022scienceqa,yue2024mmmumassivemultidisciplinemultimodal,lu2023mathvista}, demonstrating strong chain-of-thought reasoning over structured visual inputs. Furthermore, LLMs exhibit strong capabilities in
structured extraction from natural language, as demonstrated
by scene graph generation pipelines~\citep{gao2024graphdreamercompositional3dscene}
and spatial reasoning systems~\citep{li2023gligen}.
Crucially, chain-of-thought prompting~\citep{wei2022cot,kojima2022zeroshotcot}
enables LLMs to decompose physics problems into typed
entities and relationships --- objects, surfaces, physical
actions, forces, and geometric constraints --- in a form
that is amenable to downstream constraint solving. We leverage this capacity in the opposite direction: rather than parsing a diagram to solve a problem, we parse a problem to construct a diagram.

We propose \textbf{PhyDrawGen}, a pipeline that separates the semantic task of understanding a physics problem from the symbolic task of satisfying its physical constraints by employing an LLM as a structured scene graph extractor and a deterministic constraint solver as an exact physical verifier. The constraint solver produces a Planar Straight-Line Graph~(PSLG) in which physical laws are encoded as typed geometric primitives that covers mechanics, optics, and electromagnetism under a single representational framework. The complete output is a standard physics diagram comprising a full scene with labeled force arrows and per-object free-body diagrams.

This approach addresses the discrepancy between probabilistic visual representations and structural physical laws. By transforming text into symbolic blueprints prior to rendering, we ensure semantic flexibility does not compromise geometric or physical truth.

We make three contributions:

\begin{enumerate}

\item \textbf{The PhyDrawGen scene graph schema.}
A typed heterogeneous graph whose \textsc{Constraint}
node vocabulary maps directly to algebraic physical
conditions. The schema covers mechanics, optics, and electromagnetism under a unified typed vocabulary of six node classes and six edge relations.

\item \textbf{The PSLG Constraint Solver.}
A deterministic analytical solver that converts the scene
graph into a Planar Straight-Line Graph encoding force
balance, optical ray consistency, and field line topology
as typed geometric constraint primitives.

\item \textbf{A Learned Constraint Correction Loop.}
A vision-language model fine-tuned via supervised learning
on automatically generated constraint-violation examples,
implementing a propose-verify correction loop that combines
exact symbolic constraint checking with visually-grounded
language model correction.


\end{enumerate}

\section{Related Work}

\paragraph{Controlled synthesis and structured generation.}
While diffusion models and spatial adapters have revolutionized text-to-image synthesis~\citep{rombach2022ldm, ramesh2022dalle2, saharia2022photorealistictexttoimagediffusionmodels, ho2020ddpm, song2021ddim, zhang2023controlnet, mou2023t2iadapter, ye2023ipadapter}, they inherently optimize for visual plausibility under a learned prior rather than hard symbolic constraints. Consequently, even with advanced layout grounding~\citep{li2023gligen, bar2023multidiffusion} or sketch guidance~\citep{xing2026diffsketchertextguidedvector, vinker2022clipasso}, these architectures cannot guarantee that generated force arrows or optical rays adhere to Newton's or Snell's laws. To enforce physical coherence, recent work has increasingly turned to structured intermediate representations. By conditioning synthesis on typed relational graphs~\citep{johnson2018sg2im, xu2017sgg, gao2024graphdreamercompositional3dscene} and utilizing neuro-symbolic frameworks that decouple semantic reasoning from deterministic verification~\citep{huang2026learn2foldstructuredorigamigeneration}, systems can achieve robust constraint satisfaction. PhyDrawGen extends this paradigm; we leverage LLM chain-of-thought spatial reasoning~\citep{li2023gligen, openai2023gpt4v, liu2023llava, bai2023qwenvl, wei2022cot, kojima2022zeroshotcot} but replace general spatial semantics with a domain-specific vocabulary grounded strictly in classical physics.

\begin{figure*}[h]
    \centering
    \includegraphics[width=1\linewidth]{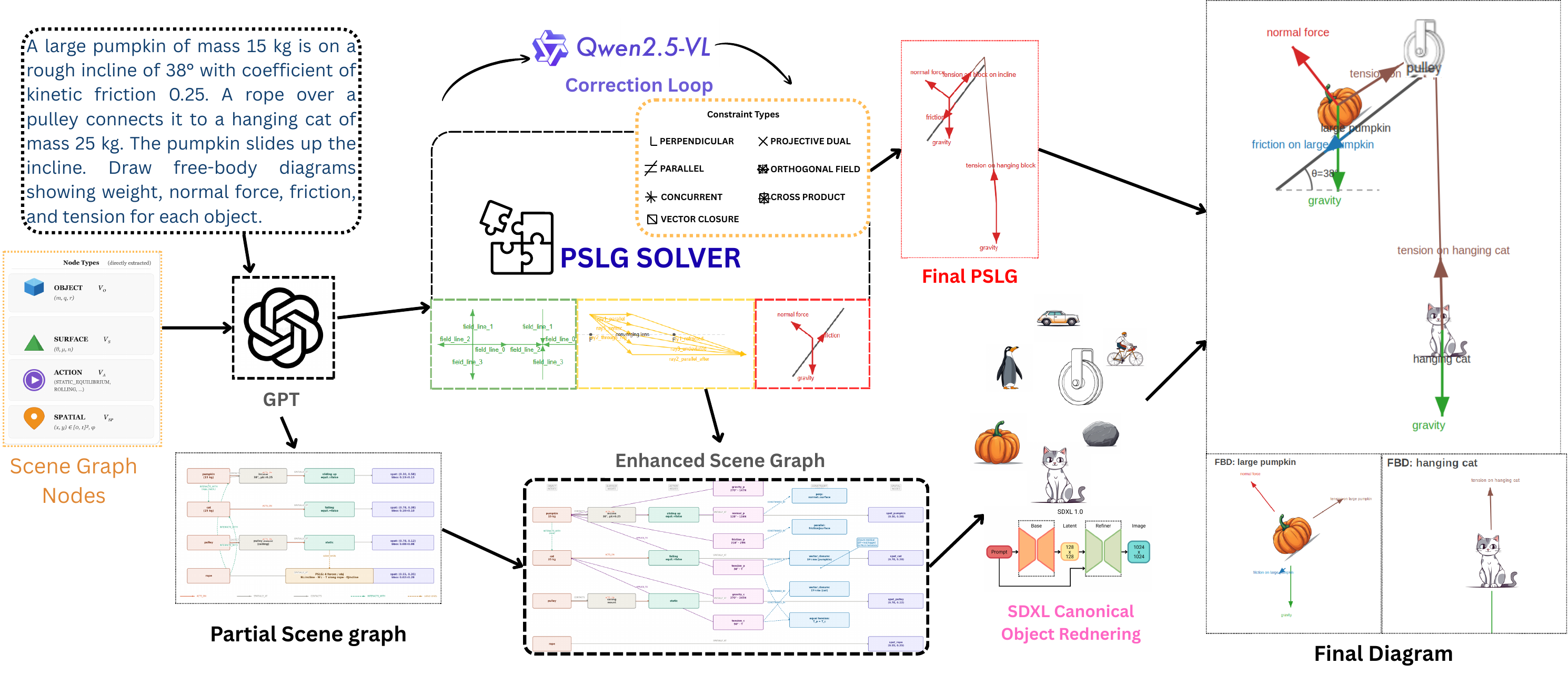}
    \caption{The overall pipeline of PhyDrawGen. To bridge semantic understanding and algebraic exactness, a language model (GPT-4o) extracts a typed scene graph from the text (left box). A solver converts it into a Planar Straight-Line Graph enforcing physical primitives (middle box), while a Qwen-VL loop iteratively corrects violations before rendering.}
    \label{fig:placeholder}
\end{figure*}

\paragraph{Physics reasoning and diagram understanding.}
The ability of vision-language models to reason over physics diagrams has been extensively benchmarked~\citep{he2024olympiadbench,xiang2025seephysdoesseeinghelp, lu2022scienceqa,yue2024mmmumassivemultidisciplinemultimodal,lu2023mathvista}, revealing strong chain-of-thought performance on diagram-based problem solving across mechanics, optics, and electromagnetism. These benchmarks establish that current models can \emph{interpret} physics diagrams with high accuracy, yet the inverse capability which is \emph{generating} physically correct diagrams from natural language has received minor  attention. Physics education research has long documented that even advanced students who conceptually understand force and motion frequently fail to construct geometrically correct diagrams~\citep{Vignal_2022, hestenes1992fci}. Translating conceptual understanding into geometric exactness requires a formal mathematical foundation. We draw on constrained quasiconformal mapping~\citep{lai2026optimizationconstrainedquasiconformalmapping} and flat-foldability theory~\citep{demaine2007geometric, 10.5555/313852.313918, hull2002combinatorics}, which establish a theoretical link between origami vertex closure and force balance, to build our PSLG constraint primitives. By integrating this rigorous geometric vocabulary, PhyDrawGen provides the first end-to-end framework that formalizes physics diagram generation as a structured prediction problem with exact algebraic verification.
\section{Methodology}

\subsection{Scene Graph Extraction}
\label{sec:extraction}
Given a physics problem text $P$, we extract a typed heterogeneous scene graph $\mathcal{G} = (\mathcal{V}, \mathcal{E})$ using GPT-4o with chain-of-thought self-checking~\citep{wei2022cot,openai2023gpt4v}. To structure the physical semantics, the node set $\mathcal{V}$ is partitioned into six classes: \textsc{Object} $\mathcal{V}_O$ (mass, charge, radius), \textsc{Surface} $\mathcal{V}_S$ (inclination, friction, refractive index), \textsc{Action} $\mathcal{V}_A$ (e.g., \textsc{static\_equilibrium}, \textsc{rolling}), \textsc{Force} $\mathcal{V}_F$ (type, direction, normalized magnitude), \textsc{Spatial} $\mathcal{V}_{Sp}$ (normalized 2D coordinates and orientation), and \textsc{Constraint} $\mathcal{V}_C$ (explicit geometric conditions like \textsc{perpendicular} or \textsc{vector\_closure}). The edge set $\mathcal{E}$ captures relations (\textsc{Acts\_On}, \textsc{Contacts}, \textsc{Interacts\_With}, \textsc{Applies\_To}, \textsc{Spatially\_At}, \textsc{Constrained\_By}), each typed by domain $d \in \{\textsc{mech}, \textsc{optics}, \textsc{em}\}$.

The LLM generates a partial graph $\mathcal{G}_{\text{LLM}} = (\mathcal{V}_{\text{LLM}}, \mathcal{E}, \delta)$, where $\mathcal{V}_{\text{LLM}} = \mathcal{V}_O \cup \mathcal{V}_S \cup \mathcal{V}_A \cup \mathcal{V}_{Sp}$ contains only the extractable entities, and $\delta$ maps edges to their physics domains. This extraction factors autoregressively over typed nodes (type $\tau_v$, attributes $\mathbf{a}_v$) and edges $\tau_e$. $p_\theta(\mathcal{G}_{\text{LLM}} \mid P) = $:
\begin{equation}
\prod_{v \in \mathcal{V}_{\text{LLM}}} p_\theta(\tau_v, \mathbf{a}_v \mid P) \prod_{(u,v,\tau_e) \in \mathcal{E}} p_\theta(\tau_e \mid u, v, P).
\label{eq:extraction}
\end{equation}

Extraction follows a rigorous five-step prompt sequence: (A) identify typed entities; (B) classify action states; (C) map inter-object contacts; (D) infer spatial positions; and (E) self-verify kinematic consistency (e.g., ensuring all \textsc{Force} nodes have targets and derivable normals). Crucially, the force and constraint sets ($\mathcal{V}_F, \mathcal{V}_C$) are not sampled by the LLM. Instead, they are instantiated deterministically by the downstream constraint solver conditioned on $\mathcal{G}_{\text{LLM}}$. The final graph is serialized as JSON and schema-validated before rendering.

\subsection{PSLG Constraint Solver}
\label{sec:solver}

Given $\mathcal{G}_{\text{LLM}}$, the solver produces a
planar straight-line graph (PSLG)
$\mathcal{H} = (\mathcal{P}, \mathcal{A}, \mathcal{C})$
— a typed straight-line embedding in $[0,1]^2$ with
vertex set $\mathcal{P}$, directed edge set $\mathcal{A}$
carrying absolute angles $\theta_a \in [0,2\pi)$, and
geometric constraint set $\mathcal{C}$. Physical laws are encoded as typed constraint primitives, one per domain, derived analytically from the scene graph

.

\paragraph{Mechanics.}
For every \textsc{Object} node $o$ with action
\textsc{static\_equilibrium}, the solver enforces the
vector closure condition
\begin{equation}
\sum_{i} \mathbf{F}_i = \mathbf{0},
\label{eq:closure}
\end{equation}
where the sum is over all force vectors applied to $o$.
This is a \textsc{concurrent\_star} primitive: all force
edges meet at the object centroid and their vector sum
closes~\citep{demaine2007geometric}.
For a \textsc{Contacts}$(o, s)$ edge with surface
inclination $\theta$, the normal force direction is
constrained by
\begin{equation}
\hat{\mathbf{N}} \perp \hat{\mathbf{s}},\quad
\hat{\mathbf{s}} = (\cos\theta,\,\sin\theta),
\label{eq:normal}
\end{equation}
the friction force direction satisfies
$\hat{\mathbf{f}} \parallel \hat{\mathbf{s}}$ with sign
determined by the direction of impending motion, and gravity
is fixed at $\hat{\mathbf{g}} = (0, -1)$.
For \textsc{rolling} objects, the torque constraint
$\tau = f \cdot r = I\alpha$ is encoded as an additional
\textsc{Constraint} node linking the friction force edge to
the object radius.
For non-equilibrium objects, Equation~\ref{eq:closure} is
relaxed and the residual $\mathbf{F}_\text{net} =
\sum_i \mathbf{F}_i \neq \mathbf{0}$ is emitted as an
explicit net-force edge encoding the acceleration direction.

\paragraph{Optics.}
At each refracting interface, the solver enforces Snell's
law
\begin{equation}
n_1 \sin\theta_1 = n_2 \sin\theta_2,
\label{eq:snell}
\end{equation}
where $n_1, n_2$ are the refractive indices from the
\textsc{Surface} node and $\theta_1, \theta_2$ are the
angles of incidence and refraction measured from the surface
normal.
For thin lenses, the solver computes the image position from
the thin lens equation
\begin{equation}
\frac{1}{f} = \frac{1}{d_o} + \frac{1}{d_i},
\label{eq:thinlens}
\end{equation}
and encodes the incident parallel ray bundle and refracted
convergent fan as a \textsc{projective\_dual} primitive.
At mirror surfaces, the reflection law
$\theta_r = \theta_i$ is enforced as a
\textsc{Constraint} node with type
\textsc{angle\_fixed}~\citep{hecht}.

\paragraph{Electromagnetism.}
For point charge interactions, the force on charge $q_A$
due to charge $q_B$ separated by displacement $\mathbf{r}$
is directed along
\begin{equation}
\hat{\mathbf{F}}_{AB} = \text{sgn}(q_A q_B)\,
\frac{\mathbf{r}}{|\mathbf{r}|},
\label{eq:coulomb}
\end{equation}
with sign enforcing attraction for opposite charges and
repulsion for like charges~\citep{griffiths2013introduction}.
Field line edges are emitted as directed radial stars with
edge count proportional to $|q|$, subject to the planarity
constraint that no two field line edges cross, encoding
Gauss's law $\oint \mathbf{E} \cdot d\mathbf{A} =
Q_\text{enc}/\varepsilon_0$~\citep{griffiths2013introduction}.
For Lorentz force problems, the force direction satisfies
\begin{equation}
\mathbf{F} = q(\mathbf{v} \times \mathbf{B}),
\label{eq:lorentz}
\end{equation}
encoded as a \textsc{cross\_product} constraint node
requiring $\mathbf{F} \perp \mathbf{v}$ and
$\mathbf{F} \perp \mathbf{B}$
simultaneously.
For uniform fields, field line edges form a
\textsc{parallel} bundle with translational symmetry
encoding $\nabla \times \mathbf{E} = 0$.
The complete PSLG is verified against all active constraint
nodes before rendering; any violation raises a solver
exception and triggers the correction loop described in
Section~\ref{sec:sft}.

\subsection{Enriched Scene Graph and Per-Object Canonical Rendering}
\label{sec:esg-canonical}
Given the PSLG $\mathcal{H}$ from Section \ref{sec:solver}, we derive an
\emph{enriched scene graph} $\mathcal{G}_E$ from
\textsc{object\_corner} vertices of each object $o \in \mathcal{V}_O$
and recording its verified bounding box $\mathbf{b}_o$, centroid
$\mathbf{p}_o$, and contact point $\mathbf{c}_o$; this surfaces
geometric quantities the solver has already fixed, no new information
is introduced.  To render each object the SVG renderer requires a
label-specific canonical PNG that it embeds inside $\mathbf{b}_o$ at
the object's PSLG orientation.  This stage is fully label-driven and
human-free: for every label $\ell_o$ in $\mathcal{G}_E$ we strip
qualifiers (\emph{heavy}, \emph{small}, \emph{hanging}, \ldots) to a
head noun $\tilde\ell_o$, slot it into a flat-illustration template
$\phi(\tilde\ell_o)$, and draw $N$ candidates from SDXL conditioned on
that prompt, retaining the best by a connected-component filter $\{I_o^{(k)}\big\}_{k=1}^{N} \;\sim\;
  p_{\text{SDXL}}\!\left(\cdot \,\big|\, \phi(\tilde\ell_o)\right)$. The final render object then becomes:
\begin{align}
    \qquad 
  I_o^{\star} \;=\;
  \arg\max_{k}\;
    \mathbb{1}\!\big[C(I_o^{(k)}) \geq \tau\big]\, s(I_o^{(k)}),
  \label{eq:canonical}
\end{align}
where $C(\cdot)$ is the largest-connected-component foreground ratio
after \textsc{rembg} background removal, $s(\cdot)$ is a silhouette
quality score, and $\tau = 0.75$.  The selected $I_o^{\star}$ is
cached and
reused by any later problem that mentions the same noun; a small set
of primitive keys (\textsc{point\_mass}, \textsc{rope}, \textsc{wire},
\textsc{point\_charge}, \textsc{sphere}) is drawn procedurally with
\textsc{pil} instead, while every other label including unseen cases follow
Eq.~\ref{eq:canonical} on first use.

\subsection{Constraint Correction via Supervised Fine-Tuning}
\label{sec:sft}

Despite strong LLM extraction, scene graphs occasionally
contain placement errors or missing relationships that
propagate to constraint violations in the PSLG.
We address this with a Qwen2.5-VL-3B-Instruct~\citep{bai2023qwenvl}
correction model $\pi_\phi$ fine-tuned on automatically
generated constraint-violation examples, implementing a
propose-verify loop.

\paragraph{Training data generation.}
For each problem in the training set, we run the PSLG
solver and record every constraint violation with its
analytical correction.
A violation instance is a tuple
$(I_\text{svg},\, \mathcal{G},\, c,\, \Delta^*)$,
where $I_\text{svg}$ is the rendered SVG image,
$\mathcal{G}$ is the current scene graph JSON,
$c$ is a structured violation description, and
$\Delta^*$ is the analytically derived correction patch.
Correction patches are derived without human annotation:
for a force balance violation with residual
$\mathbf{r} = \sum_i \mathbf{F}_i \neq \mathbf{0}$, the
correction adjusts the responsible force angle by
$-\arg(\mathbf{r})$; for a normal direction violation, the
correction sets $\theta_N = \theta_s + 90^\circ$; for a Snell
violation, the correction recomputes $\theta_2$ from
Equation~\ref{eq:snell}.
This yields a training corpus derived entirely from the
pipeline's own failures, requiring no external annotation.


\paragraph{Training objective.}
The model is fine-tuned with LoRA
adapters~\citep{hu2021loralowrankadaptationlarge} to predict correction patches
in a fixed JSON schema, minimising $\mathcal{L}_\text{SFT}(\phi) =$
\begin{equation}
\mathbb{E}_{(I,\mathcal{G},c,\Delta^*) \sim \mathcal{D}}
\left[
-\sum_k \log \pi_\phi\!\left(\Delta^*_k \mid
I_\text{svg},\, \mathcal{G},\, c,\,
\Delta^*_{<k}\right)
\right]
\label{eq:sft}
\end{equation}
where $\Delta^*_k$ denotes the $k$-th token of the
correction patch.
The training signal is further structured by three
physics-informed auxiliary losses.
The closure loss
\begin{equation}
\mathcal{L}_\text{closure} =
\left\|\sum_{i} \hat{\mathbf{u}}_i\right\|^2,
\label{eq:lclosure}
\end{equation}
where $\hat{\mathbf{u}}_i$ is the unit vector in the
direction of the $i$-th predicted force arrow, penalises
deviations from vector closure at equilibrium vertices and
is set to zero for non-equilibrium objects.
The geometry loss
\begin{equation}
\mathcal{L}_\text{geom} =
\sum_{j} \left\|\mathbf{p}_j -
\mathbf{p}_j^*\right\|^2,
\label{eq:lgeom}
\end{equation}
penalises deviation of each predicted attachment point
$\mathbf{p}_j$ from its physics-correct target
$\mathbf{p}_j^*$ derived from the SAM-detected object
boundary.
The relation loss
\begin{equation}
\mathcal{L}_\text{rel} =
\sum_{e \in \mathcal{E}_\text{contact}}
\left(\hat{\mathbf{N}}_e \cdot \hat{\mathbf{s}}_e\right)^2,
\label{eq:lrel}
\end{equation}
penalises deviation from the perpendicularity constraint in
Equation~\ref{eq:normal} for each active contact edge.
The total training loss is
\begin{equation}
\mathcal{L} = \mathcal{L}_\text{SFT}
+ \mathcal{L}_\text{closure}
+ \mathcal{L}_\text{geom}
+ \mathcal{L}_\text{rel},
\label{eq:total}
\end{equation}

\label{sec:results}
\begin{figure*}[h]
    \centering
    \includegraphics[width=\textwidth,
    height=0.4\textheight,keepaspectratio]{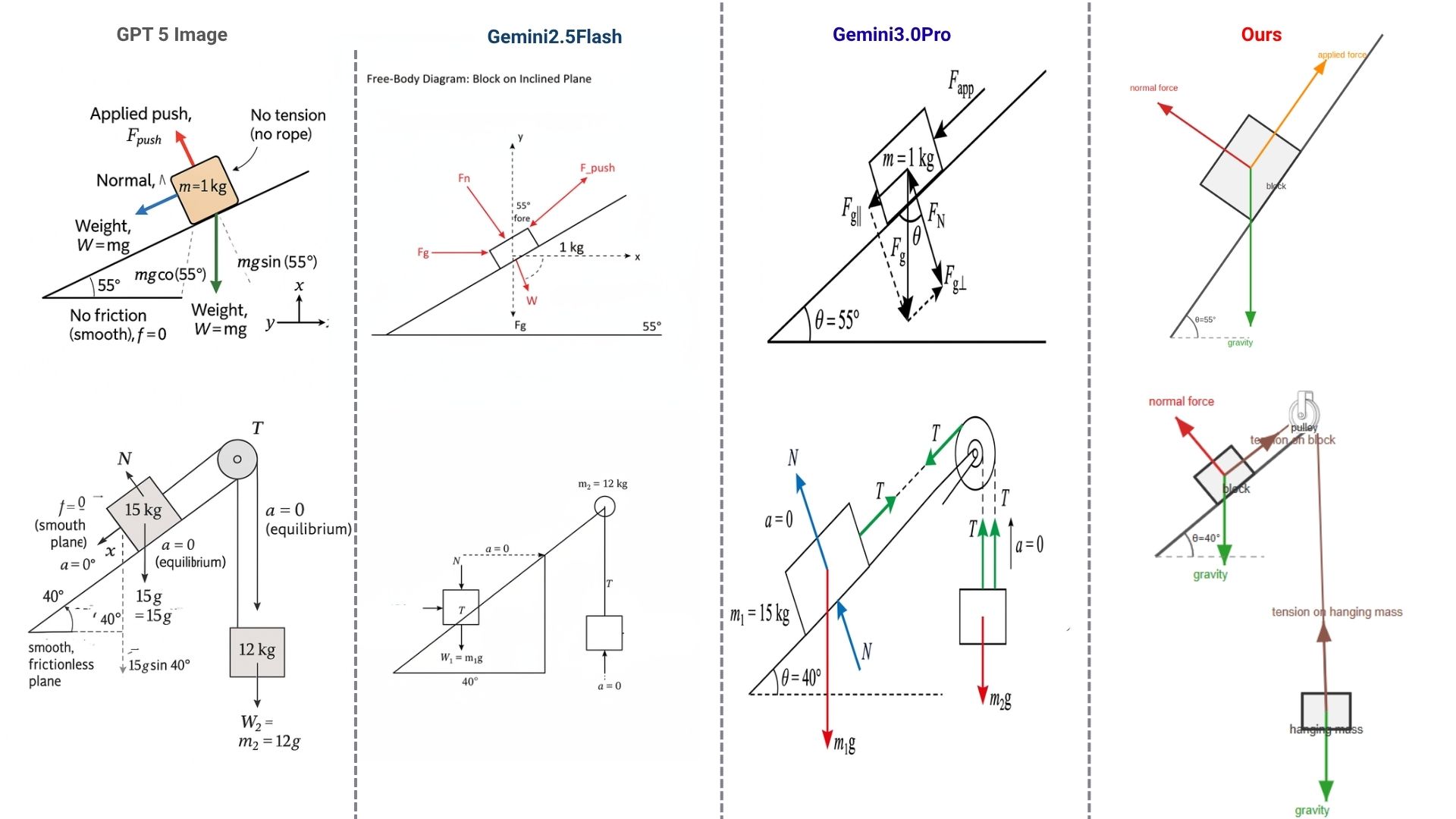}
    \caption{Qualitative comparison on standard mechanics
    problems. Each row shows the input problem, outputs
    from GPT-5-image, Gemini~2.5~Flash, Gemini~3~Pro, and
    PhyDrawGen (ours). Force arrows in baseline outputs
    frequently point in incorrect directions or violate
    equilibrium; PhyDrawGen arrows are derived analytically
    from the PSLG and satisfy all encoded constraints by
    construction.}
    \label{fig:mech_qual}
\end{figure*}

\subsection{Evaluation Metrics}
\label{sec:metrics}

We use a single set of \emph{neutral} ground-truth arrows, derived once per problem from the GPT-extracted scene graph $\mathcal{G}_\text{LLM}$ and the problem text $P$ via closed-form textbook physics formulas (catalogued in Appendix~\ref{app:neutral-gt}). This ground truth is completely independent of the constraint solver. Let $\mathcal{A}^\star(p) = \{(\theta_i^\star,\ell_i^\star)\}_{i=1}^{N_p}$ denote the neutral ground-truth set of angles and force labels for problem $p$, and let $\hat{\mathcal{A}}(p) = \{(\hat\theta_j,\hat\ell_j)\}_{j=1}^{\hat N_p}$ denote the predicted set (Hough segments or VLM-judged arrows, defined below). All three metrics use this same neutral $\mathcal{A}^\star(p)$ and are scored at a tolerance threshold of $\tau=10^\circ$ unless otherwise noted.

\paragraph{Hough-CSR.}
For each rendered diagram, we apply a Canny edge detector followed by the probabilistic \textsc{HoughLinesP} segment detector~\citep{MATAS2000119} to produce $\hat{\mathcal{A}}(p)$. A greedy minimum-angular-distance assignment matches each predicted segment to one ground-truth arrow within a $20^\circ$ admissibility gate, yielding a matched set $\mathcal{M}(p)$. The binary \textbf{Hough-CSR} (Constraint Satisfaction Rate) metric for a problem is 1 if and only if every expected arrow is matched ($|\mathcal{M}(p)| = N_p$) and the maximum angular error among all matches is $\le \tau$. We additionally report the mean angular error and the detection rate $|\mathcal{M}(p)|/N_p$ averaged across problems.

\paragraph{VCSR and LblCSR.}
A frozen VLM judge (Claude Opus 4.7) receives the rendered diagram alongside the problem text and is prompted to enumerate every visible arrow as a tuple $(\hat\theta_j, \hat\ell_j, \hat{o}_j)$ of angle, force-type label, and originating object. The judge's output is parsed into $\hat{\mathcal{A}}(p)$ and Hungarian-matched~\citep{Kuhn1955Hungarian} to the neutral ground truth $\mathcal{A}^\star(p)$ under a composite cost $c_{ij} = \alpha\Delta\theta_{ij} + \gamma\mathbb{1}[\hat\ell_j \neq \ell_i^\star]$. We report two binary problem-level metrics: the angle-only \textbf{VCSR}, which requires the maximum angular error of all matches to be $\le \tau$, and the stricter \textbf{LblCSR}, which additionally requires exact label agreement ($\hat\ell_j = \ell_i^\star$) for all matched pairs. Auxiliary statistics include the mean angular error and the label-match rate.

\paragraph{Blind judge.}
To isolate human-perceptible diagram quality, a separate frozen VLM (Claude Sonnet 4.6) is shown only the rendered diagram and the problem text $P$. It is prompted to evaluate six force categories $\mathcal{F}$=\{\textsc{gravity}, \textsc{normal}, \textsc{friction}, \textsc{tension}, \textsc{spring}, \textsc{applied}\} assigning each a verdict of \textsc{correct}, \textsc{wrong}, \textsc{missing}, or \textsc{n/a}. No solver output or ground truth is provided. The \textbf{Blind score} for a problem is simply the fraction of \textsc{correct} verdicts out of all applicable categories (excluding \textsc{n/a}). We report the unweighted mean across the problem set, with per-category correctness rates broken out separately (see Appendix~\ref{app:blind-judge} for the exact prompt and inter-judge sanity checks).

\section{Experiments and Results}

\subsection{Experimental Setup}

\paragraph{Implementation details.}
The full PhyDrawGen pipeline runs on a single NVIDIA RTX
4090 32\,GB GPU.
Scene graph extraction uses GPT-4o via API with
temperature~0.
The PSLG constraint solver and SVG renderer are
deterministic and run in under one second per problem.
The Qwen2.5-VL-3B-Instruct correction model is fine-tuned with
LoRA adapters~\citep{hu2021loralowrankadaptationlarge} for 16K iterations on the
automatically generated violation dataset of 1.8k instances.
Object rendering uses SDXL with MistoLine
ControlNet~\citep{zhang2023controlnet} at
1024$\times$1024 resolution.
Training the correction model requires 24--27\,GB;
inference for a complete diagram requires 6--9\,GB.

\paragraph{Baselines.}
We compare against three state-of-the-art generative
models: GPT-5-image, Gemini~2.5~Flash, and Gemini~3~Pro.
All three are prompted directly with the problem text and
asked to generate a physics diagram with labeled force
arrows; no additional spatial conditioning or structured
intermediate representation is provided.

\subsection{Standard Textbook Problems}
\begin{table}[h]
\centering
\small
\setlength{\tabcolsep}{3.5pt}
\begin{tabular}{lccccc}
\toprule
\textbf{Method} & \textbf{H-V1} &
\textbf{VCSR} & \textbf{LblCSR} & $\Delta_\text{ang}$
& \textbf{Blind} \\
\midrule
GPT-5-image & 78.9\% & 79.7\% & 47.1\% &  $2.2^\circ$ & 49.8\% \\
Gemini~2.5~Flash & 68.4\% & 73.7\% & 31.8\% &  $3.0^\circ$ & 33.3\% \\
Gemini~3~Pro & 89.5\% & 57.9\% & 41.2\% & $3.6^\circ$ & 60.2\% \\
\midrule
\textbf{Ours} & \textbf{78.9\%} & \textbf{94.7\%} &
\textbf{77.9\%} & \textbf{$0.4^\circ$} & \textbf{65.8\%} \\
\bottomrule
\end{tabular}
\caption{Results on mechanics, optics,
and electromagnetism problems. VCSR is the VLM-judged
constraint satisfaction rate measuring geometric arrow
correctness; LblCSR additionally requires correct force
labels; $\Delta_\text{ang}$ is mean angular error in
degrees; Blind is the blind VLM judge score.
PhyDrawGen achieves the lowest angular error by 5x and the highest scores on all semantic and
holistic metrics.}
\label{tab:mech19}
\end{table}

\begin{table}[h]
\centering
\small
\setlength{\tabcolsep}{3.5pt}
\begin{tabular}{lcccccc}
\toprule
\textbf{Method} & \textbf{Grv} & \textbf{Nrm} &
\textbf{Frc} & \textbf{Tns} & \textbf{Spr} &
\textbf{Apl} \\
\midrule
GPT-5-image      & 88\%  & 65\%  & 38\%  & 86\%  & 21\%  & 100\% \\
Gemini~2.5~Flash & 72\%  & 78\%  & 17\%  & 35\%  & 78\%  & 33\%  \\
Gemini~3~Pro     & 100\% & 78\%  & 67\%  & 100\% & 35\%  & 60\%  \\
\midrule
\textbf{Ours}    & \textbf{100\%} & \textbf{90\%} &
\textbf{86\%} & 60\% & \textbf{100\%} & 70\% \\
\bottomrule
\end{tabular}
\caption{Per-force-type blind judge scores
mechanics subset. Grv: gravity. Nrm: normal. Frc:
friction. Tns: tension. Spr: spring. Apl: applied.
PhyDrawGen leads on four of six force types.}
\label{tab:mech19_force}
\end{table}
Table~\ref{tab:mech19} reports results on problems across
mechanics, optics, and electromagnetism problems.
PhyDrawGen achieves 94.7\% VCSR and 77.9\% LblCSR,
outperforming all baselines on both correctness metrics
and reaching the highest blind judge score of 65.8\%,
above Gemini~3~Pro at 60.2\%.
The mean angular error of 0.4° is an order of magnitude
lower than every baseline, confirming that PSLG-derived
arrows are geometrically exact rather than approximately
correct.

\begin{figure*}[t!]
    \centering
    \includegraphics[width=\textwidth,
    height=0.4\textheight,keepaspectratio]{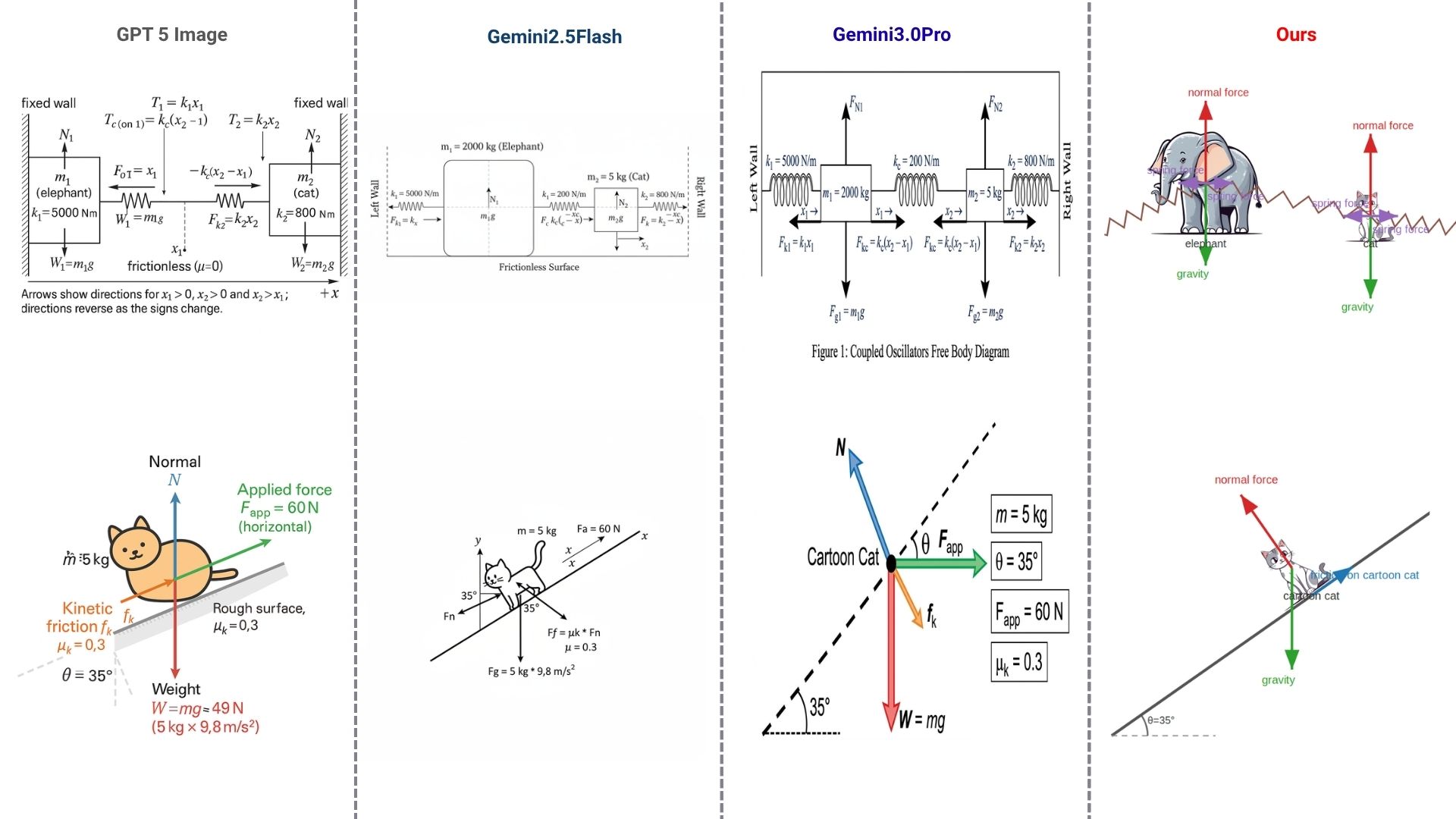}
    \includegraphics[width=\textwidth,
    height=0.4\textheight,keepaspectratio]{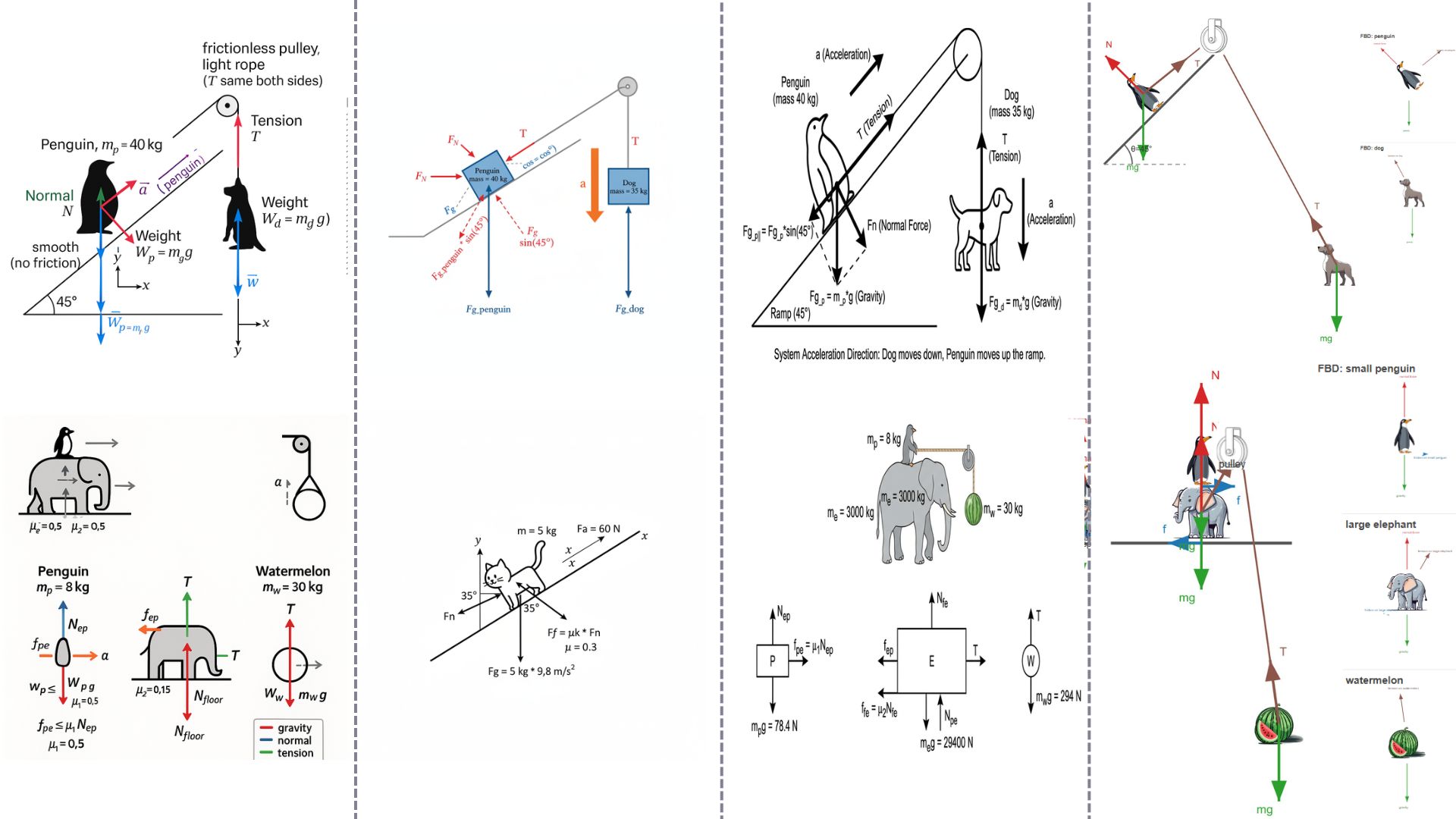}
    \caption{Qualitative comparison on open-vocabulary
    problems. Baseline models produce visually plausible
    renderings of unusual objects but systematically
    misplace or omit force arrows. PhyDrawGen derives all
    arrow geometry from the PSLG independently of object
    identity, achieving correct force placement for
    elephants, cyclists, and penguins as readily as for
    standard physics blocks.}
    \label{fig:ov_qual}
\end{figure*}

Table~\ref{tab:mech19_force} breaks down blind judge
scores by force type on the mechanics subset.
PhyDrawGen achieves 100\% on gravity and spring forces
and leads on friction, consistent with the PSLG's
explicit enforcement of vertical gravity, surface-
perpendicular normals, and surface-parallel friction
directions.
Tension and applied force scores are lower across all
methods.

\subsection{Open-Vocabulary Problems}

\begin{table}[h]
\centering
\small
\setlength{\tabcolsep}{3.5pt}
\begin{tabular}{lccccc}
\toprule
\textbf{Method} & \textbf{H-V1} &
\textbf{VCSR} & \textbf{Lbl-CSR} & $\Delta_\text{ang}$
& \textbf{Blind} \\
\midrule
GPT-5-image      & 72.6\% & 46.2\% & 23.1\% & $4^\circ$ & 41.7\% \\
Gemini~2.5~Flash & 69.2\% & 61.5\% & 15.4\% & $3.6^\circ$ & 33.9\% \\
Gemini~3~Pro     & 76.9\% & 53.8\% & 23.1\% & $2.7^\circ$ & \textbf{55.2\%} \\
\midrule
\textbf{Ours}    & \textbf{76.9\%} & \textbf{92.3\%} &
\textbf{73.8\%} & \textbf{$0.7^\circ$} & 53.6\% \\
\bottomrule
\end{tabular}
\caption{Results on Open-Vocabulary-65, where physical
objects are arbitrary real-world entities such as elephants,
dogs, penguins, pumpkins, cyclists, rather than
standard physics apparatus.
PhyDrawGen leads on all objective metrics by a wide margin,
demonstrating that constraint satisfaction generalises
when object identity provides no implicit geometric prior.
The blind judge gap versus Gemini~3~Pro reflects the
latter's stronger visual rendering of unusual objects.}
\label{tab:unusual13}
\end{table}
Open-vocabulary problems probe a qualitatively different
capability from standard textbook diagrams: the pipeline
must generalise to objects e.g., elephants, watermelons,
penguins, cartoon cats, cyclists  that carry no
implicit geometric prior about where forces should attach
or in what direction they should point.
A block on an incline always has a rectangular footprint;
an elephant on the same incline does not, and no
physics-specific training signal tells the model what
an elephant looks like.
This set therefore directly tests whether physical
constraint satisfaction is genuinely decoupled from
object identity, which is the central architectural
claim of PhyDrawGen.

Figure~\ref{fig:ov_qual} shows qualitative outputs on
three representative open-vocabulary problems;
Table~\ref{tab:unusual13} reports quantitative results
across the full set.
PhyDrawGen achieves 92.3\% VCSR against GPT-5-image's
46.2\%, a margin of more than 46 percentage points,
and Label-CSR of 73.8\% leads all baselines by a
similarly large margin, confirming that force directions
are derived from action and contact edges in the scene
graph rather than from object identity.
The blind judge score of 53.4\% marginally lags
Gemini~3~Pro's 55.2\%, a notable result given that
Gemini produces visually richer scene synthesis. The objective correctness advantage is preserved even
where our visual rendering of unusual objects is more
schematic.

\section{Ablation Studies}
\label{sec:ablation}
\begin{table}[h]
\centering
\small
\setlength{\tabcolsep}{3.5pt}
\begin{tabular}{lccccc}
\toprule
\textbf{Condition} & \textbf{H-V1} &
\textbf{VCSR} & \textbf{LblCSR} & $\Delta_\text{ang}$ &
\textbf{Conv.} \\
\midrule
Clean GT (ceiling)       & 33.3\% & 68.0\% & 50.0\% & $1.6^\circ$ & ---  \\
Perturbed (no SFT)       & 23.8\% & 48.8\% & 32.5\% & $5^\circ$ & ---  \\
\textbf{Perturbed + SFT} & {32.8} & {61.7} & {50.0} & {$1.2^\circ$} & {78}\% \\
\bottomrule
\end{tabular}
\caption{Ablation: SFT recovery on 80 perturbed instances.
\textbf{Conv.}\ is the fraction fully resolved within $T_\text{max}{=}5$
iterations. VCSR is the primary signal; Hough-V1 is noisy on this
small mechanics-only set.}
\label{tab:ablation-sft}
\end{table}
To evaluate the contribution of the Qwen-VL constraint-correction
loop (Section \ref{sec:sft}), we construct a synthetic perturbation testbed
of 80 held-out mechanics instances by applying two violation classes
to clean GPT-4o-extracted scene graphs: \textsc{Normal-Direction
Error} (incline surface angle rotated by $\pm[12^\circ,90^\circ]$)
and \textsc{BBox-Placement Error} (spatial node position shifted by
$\pm[0.04,0.25]$ on a random axis).
We compare three conditions scored against the same neutral ground
truth: \textbf{Clean GT} (unperturbed ceiling), \textbf{Perturbed
(no SFT)} (perturbed scene graph rendered as-is), and
\textbf{Perturbed + SFT} (perturbed graph corrected by the
propose-verify loop with $T_\text{max}{=}5$).
As reported in Table~\ref{tab:ablation-sft}, removing the SFT loop
drops VCSR by 19.2\,pp (68.0\%~$\to$~48.8\%) and raises mean angular
error from $1.6^\circ$to $5^\circ$, confirming that perturbations propagate
faithfully through the solver to the rendered diagram.
The full pipeline with SFT recovers 12.9\,pp of that gap, with
78\% of instances fully converging within $T_\text{max}$ iterations.



\section{Conclusion}
\label{sec:conclusion}
This paper proposes PhyDrawGen, a neuro-symbolic pipeline for generating physically rigorous science diagrams from natural language. By decoupling semantic scene understanding from physical constraint satisfaction, PhyDrawGen leverages an LLM to extract typed scene graphs and a deterministic solver to enforce exact geometric primitives (force balance, optical paths, and field topologies) via a Planar Straight-Line Graph (PSLG). We demonstrate through the introduction of a 1,449-problem benchmark that this architecture effectively mitigates systematic hallucinations, misplaced force vectors, and conservation law violations prevalent in purely diffusion-based models. 
\section*{Limitations}
\label{sec:limitations}
While PhyDrawGen successfully enforces hard physical constraints, its reliance on a deterministic, 2D Planar Straight-Line Graph (PSLG) restricts the framework to classical planar interactions, presenting an opportunity for future work to explore learned or dynamic constraint solvers capable of handling arbitrary 3D topologies. Furthermore, our evaluation benchmark is curated primarily from standard early-undergraduate physics curricula, omitting highly complex or abstract domains such as Olympiad-level mechanics or quantum phenomena (e.g., Feynman diagrams). Our baseline comparisons focus exclusively on state-of-the-art proprietary models (such as GPT-5-image and Gemini 3 Pro) because current open-weight generative models lack the zero-shot compositional text-to-image capabilities required to serve as meaningful, apples-to-apples baselines for this specific diagram-generation task. Finally, because the pipeline heavily depends on an initial LLM extraction, severely underspecified geometric parameters in the problem text can result in structural omissions that the downstream visual correction loop capped at $T_\text{max}{=}5$ iterations cannot always resolve.

\section*{Ethics Statement}
PhyDrawGen is intended to support physics education and the authoring of scientifically accurate teaching material; it generates diagrams from textbook-style problem statements and does not process personal, sensitive, or private data. Our evaluation benchmark is constructed from standard early-undergraduate physics problems and contains no human-subject information. The human-judge study reported in Appendix~\ref{app:human-judge} involved adult volunteers who participated voluntarily and anonymously, viewing only rendered diagrams and the accompanying problem text; no personal data was collected. The pipeline relies on a combination of proprietary APIs (e.g., GPT-4o) and open-weight models (Qwen2.5-VL, SDXL), each used in accordance with its respective terms of use. Finally, although our constraint solver enforces physical correctness by construction, generated diagrams may still contain residual errors; we therefore caution against treating the system as a sole authority in safety-critical or assessment settings without expert review.

\section*{Acknowledgements}

We thank Md Taif Islam Tonmoy and Muhammad Jiyad Hasan for their contributions to formulating the physical reasoning underlying this work, as well as for many valuable discussions on the representation and description of physics diagrams. Their feedback on the correctness of the mechanics, optics, and electromagnetism constructions was instrumental in shaping this paper.

\section*{Generative AI Usage}
We used a generative AI assistant solely to aid manuscript writing, specifically for grammatical correction and minor stylistic polishing of author-written text. It was not used to generate research ideas, conduct experiments, or analyze results.

\bibliography{anthology,custom}
\bibliographystyle{acl_natbib}

\appendix

\section{Additional Implementation Details}
\label{sec:appendix-impl}

\subsection{GPT Scene Graph Extraction}
\label{app:extraction-prompt}

Stage 1 of the pipeline (Section \ref{sec:extraction}) is implemented as a
single GPT-4o call routed through OpenRouter with temperature~0.
The model receives a system prompt that fixes the schema (six node
classes, six edge classes, force / constraint node types reserved
for the solver), a per-object bounding-box \textsc{OBJECT-SIZE
GUIDE} with concrete anchors for unusual entities (elephant
$\approx 0.30 \times 0.22$, cat / dog $\approx 0.10 \times 0.10$,
wall $\approx 0.05 \times 0.60$, etc.), a five-step
chain-of-thought scaffold (Steps A--E covering entities, actions,
relationships, spatial layout, observed elements), and a
\emph{mandatory} self-check block that requires the model to
re-confirm every object has both a \textsc{SPATIALLY\_AT} and an
\textsc{ACTS\_ON} edge before emitting JSON. The model writes the
self-check in plain text before the final JSON fence, providing an
audit trail. We keep the prompt frozen across all main-paper
experiments. The full system prompt and the user-turn instruction
sent alongside each problem are reproduced verbatim below; a
controlled ablation that strips the chain-of-thought scaffold and
self-check block while preserving the schema and critical rules
is reported in Section \ref{app:cot-naive}.

\noindent\textit{The full system prompt is split below into six
colour-coded sections for readability: the top-level schema
(blue), the four node types (green / orange / red / yellow),
and the procedure + rules (grey). All content is reproduced
verbatim; the section
headings here mirror the headings in the source.}
\vspace{0.4em}

\begin{pblue}{1. Top-level schema and edge types}
\begin{footnotesize}
\begin{flushleft}
You are a physics scene graph extractor. Output a typed JSON scene graph
describing a physics problem. The solver fills in force / constraint
nodes downstream — you do NOT emit them.

Top-level shape:
{
  "domain": ["mechanics" | "optics" | "electrostatics" | "magnetic" | "em_uniform"],
  "nodes": [...],                 // object / surface / action / spatial nodes only
  "edges": [...],                 // typed directed edges, see below
  "force_mapping_output": [],     // ALWAYS empty — solver fills
  "constraint_set":       [],     // ALWAYS empty — solver fills
  "observed_elements": {...}      // arrows / labels visible in image (image-mode only)
}

Each edge: {"source": id, "target": id, "edge_type": TYPE, "properties": {}}
  ACTS_ON         action  → object       action applies to object
  CONTACTS        object  → surface      object rests on / touches surface
  INTERACTS_WITH  object  → object       field or mechanical link
                                         (properties.interaction_type:
                                          electrostatic | gravitational |
                                          magnetic | optical | mechanical)
  SPATIALLY_AT    object | surface → spatial    locates the node
  APPLIES_TO      only for externally specified forces (rare)
\end{flushleft}
\end{footnotesize}
\end{pblue}
\begin{pgreen}{2. Object nodes (type = "object")}
{
  "id":            "obj_1",
  "type":          "object",
  "label":         "<descriptive>",
  "physical_type": ONE OF [
                     "rigid_body", "point_mass", "point_charge",
                     "lens", "mirror", "prism", "ray_source",
                     "wave_source", "screen",
                     "wire", "solenoid", "coil",
                     "pulley", "rope", "spring"
                   ],
  "mass":          float | null,
  "charge":        float | null,
  "radius":        float | null,
  "focal_length":  float | null
}

Default to "rigid_body" if no specific type fits. Pulley / rope / spring
have dedicated relationship rules — see the grey "Critical rules" box.
\end{pgreen}

\begin{porange}{3. Surface nodes (type = "surface")}
{
  "id":               "surf_1",
  "type":             "surface",
  "label":            "<descriptive>",
  "surface_type":     ONE OF ["incline", "flat", "curved", "interface", "mirror"],
  "angle_deg":        float | null,      // incline angle from horizontal
  "friction_coeff":   float | null,
  "normal_direction": [nx, ny] | null    // unit vector, only if angle missing
}

ALWAYS create a surface node for every floor / wall / ceiling / incline /
rope-anchor / optical interface visible. The force solver needs them to
derive normal and friction directions.
\end{porange}

\begin{pred}{4. Action nodes (type = "action")}
{
  "id":               "act_1",
  "type":             "action",
  "label":            "<descriptive>",
  "action_type":      ONE OF [
                        "static_equilibrium", "sliding", "rolling", "falling",
                        "orbiting", "oscillating", "flowing",
                        "refracting", "reflecting", "interfering",
                        "circular_motion", "rotating", "conducting"
                      ],
  "equilibrium":      true | false,
  "motion_direction": [vx, vy] | null
}

Read motion cues BEFORE picking action_type — do NOT default to static:
  acceleration arrow / "find a" / "a=?"   → equilibrium=false, sliding/falling
  velocity arrow shown                    → equilibrium=false
  single unopposed force                  → equilibrium=false
  spring + mass, no motion arrows         → oscillating, equilibrium=true
  balanced figure / "STATICS"             → static_equilibrium, equilibrium=true
Only use static_equilibrium when you are CONFIDENT nothing accelerates.
\end{pred}

\begin{pyellow}{5. Spatial nodes + OBJECT-SIZE GUIDE}
{
  "id":               "spat_1",
  "type":             "spatial",
  "label":            "<what this locates>",
  "position":         [x, y],         // [0,0]=top-left, [1,1]=bottom-right
  "orientation_deg":  float | null,
  "bounding_region":  [[x0,y0],[x1,y1]] | null,
  "path_type":        "circular" | "parabolic" | "linear" | "arc" | null,
  "path_params":      {} | null
}

Size anchors (normalized w × h):
  large creature / vehicle (elephant, car, horse) ...... 0.30 × 0.22
  medium creature / person (cyclist, dog, penguin) ..... 0.10 × 0.20
  small creature (cat, monkey, baby animal) ............ 0.10 × 0.10
  pulley ..........................0.08 × 0.08
  spring (zigzag span) .................. 0.20 × 0.04
  lens / mirror / prism (optics) ............ 0.06 × 0.20
  point charge / coil ............. 0.04 × 0.04


\end{pyellow}

\begin{pgray}{6. Extraction procedure, critical rules, self-check}
PROCEDURE (do internally before writing JSON):
  A. ENTITIES        identify every object + surface; assign physical_type.
  B. ACTIONS         read motion cues first; pick action_type accordingly.
  C. RELATIONSHIPS   add CONTACTS + INTERACTS_WITH edges — these drive forces.
  D. SPATIAL LAYOUT  infer positions from language: floor → y≈0.75,
                     hanging → y≈0.35, wall on left → x≈0.10.
  E. OBSERVED        (image mode) record force/velocity arrows, dimension
                     and angle labels exactly as drawn.

CRITICAL RULES:
  - EVERY object MUST have BOTH a SPATIALLY_AT edge AND an ACTS_ON edge,
    including passive objects (pulley, rope, spring, second mass).
  - rope / spring → INTERACTS_WITH (not CONTACTS) to BOTH connected objects.
    interaction_type = "mechanical".
  - Pulley: physical_type "pulley" with a CONTACTS edge to its mounting surface.
  - Use null for any numeric field you cannot determine.

SELF-CHECK (mandatory) before emitting JSON:
  A. List every object node ID.
  B. For each ID, confirm BOTH a SPATIALLY_AT and an ACTS_ON edge exist;
     if missing, add them. Passive objects are NOT exempt.
  C. For each SPATIALLY_AT edge, confirm the target spatial node exists.
  D. Confirm no node has type "force" or "constraint".
  E. Confirm nodes[] is non-empty.

\end{pgray}

\begin{pblue}{GPT-4o user-turn instruction (text-only problems)}
Read this physics problem. Work through the self-check steps (A–E) in plain text, then output the final scene graph JSON inside a ```json ... ``` fence.
Since there is no image, infer spatial positions from the semantic description. Coordinate convention: [0,0]=top-left, [1,1]=bottom-right, y increases downward.
Placement rules: object on floor/table → y≈0.75; wall on left → x≈0.10; object hanging from ceiling → y≈0.35; spring/rope anchor on wall at height h above floor → place anchor spatial node at [wall_x, floor_y - h_fraction] so it appears above the floor. Objects side by side → spread x from 0.2 to 0.8. For a block-spring-wall system: wall at x≈0.10, block at x≈0.65, both at floor y≈0.75. CRITICAL — spring SPATIALLY_AT placement:   (a) HORIZONTAL spring (spring parallel to floor, attaches at floor level): spring SPATIALLY_AT = [0.10, 0.75] (same y as floor).   (b) ANGLED spring (spring attaches to wall ABOVE floor, at height h): spring SPATIALLY_AT = [0.10, 0.50] (y = mid-height). Never use [0.10, 0.50] for a horizontal spring — that makes it render angled.
Rules: (1) nodes[] must NOT be empty — identify every object, surface, action, and spatial node from the problem text. (2) Every node ID used in edges must exist in nodes[]. (3) Leave force_mapping_output and constraint_set as empty arrays.
\end{pblue}

\subsection{Extracted Scene Graph Examples}
\label{app:sg-examples}

To make the extracted scene graph schema concrete, we show
three representative outputs, one per domain, alongside
the problem text and the scene graph visualisation rendered
from the on-disk JSON.

\begin{figure}[h]
    \centering
    \includegraphics[width=\columnwidth]{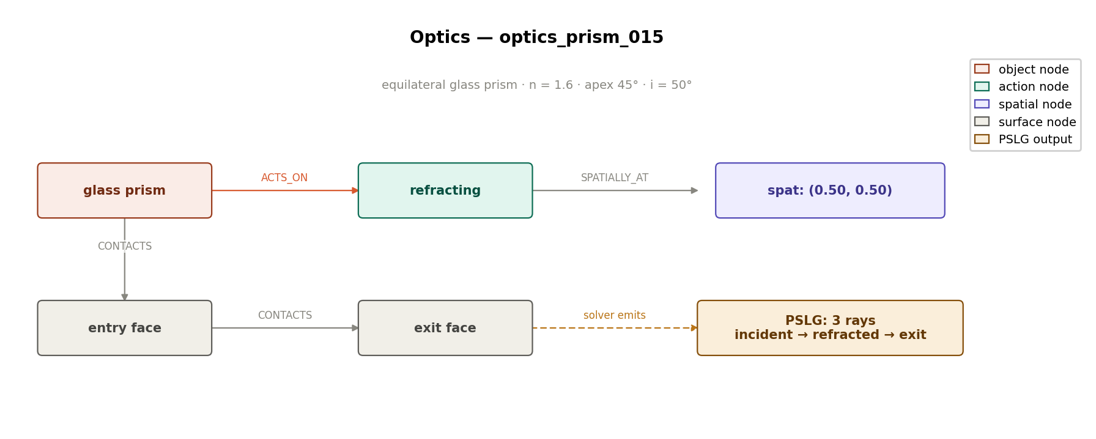}
    \includegraphics[width=\columnwidth]{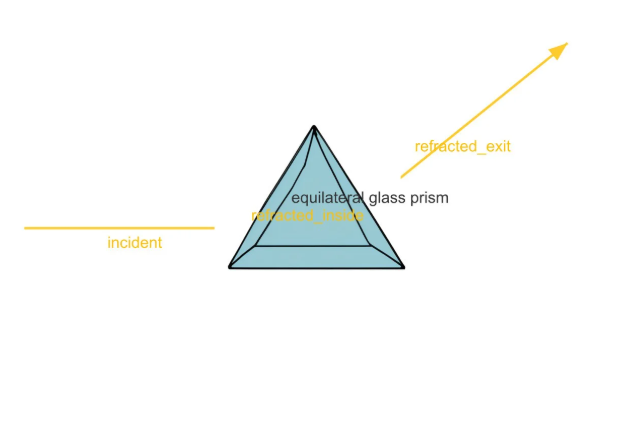}
    \caption{Top Row: Scene graph of optical prism problem (\texttt{optics\_prism\_015}). Bottom Row: Final Rendered Diagram of the System}
    \label{fig:sg-mech}

\end{figure}

\paragraph{Mechanics.}
Figure~\ref{fig:sg-mech} shows the scene graph for
\texttt{mech\_stacked\_pulley\_071}: \emph{``Two blocks
stacked on a frictionless incline connected via a rope over
a pulley to a hanging mass.''} Four object nodes (block A,
block B, hanging mass, pulley) are linked by
\textsc{Acts\_On} edges to two action nodes encoding their
physical states (stacked+sliding and hanging+accelerating).
\textsc{Contacts} edges capture the block-floor and
inter-block contact; a dashed \textsc{Interacts\_With}
edge encodes the rope coupling between the hanging mass and
the pulley. Spatial nodes carry the normalised canvas
coordinates the PSLG solver uses to instantiate force
attachment points and object bounding boxes.


\begin{figure}[h]
    \centering
    \includegraphics[width=\columnwidth]{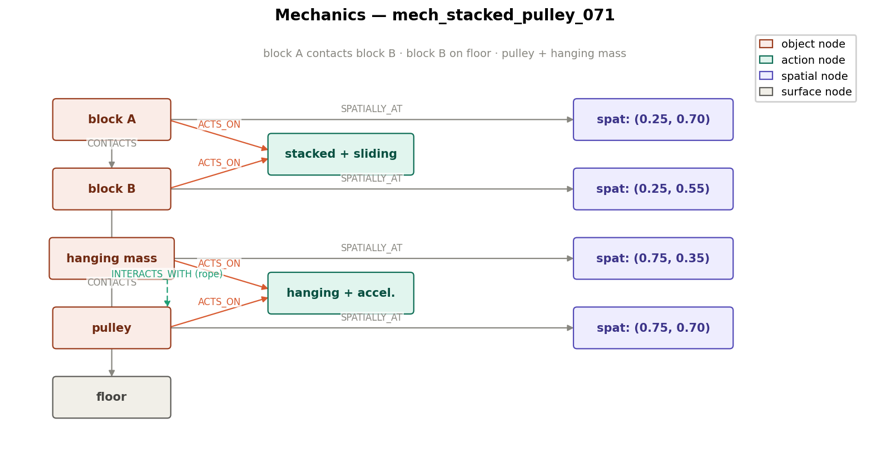}
    \includegraphics[width=\columnwidth]{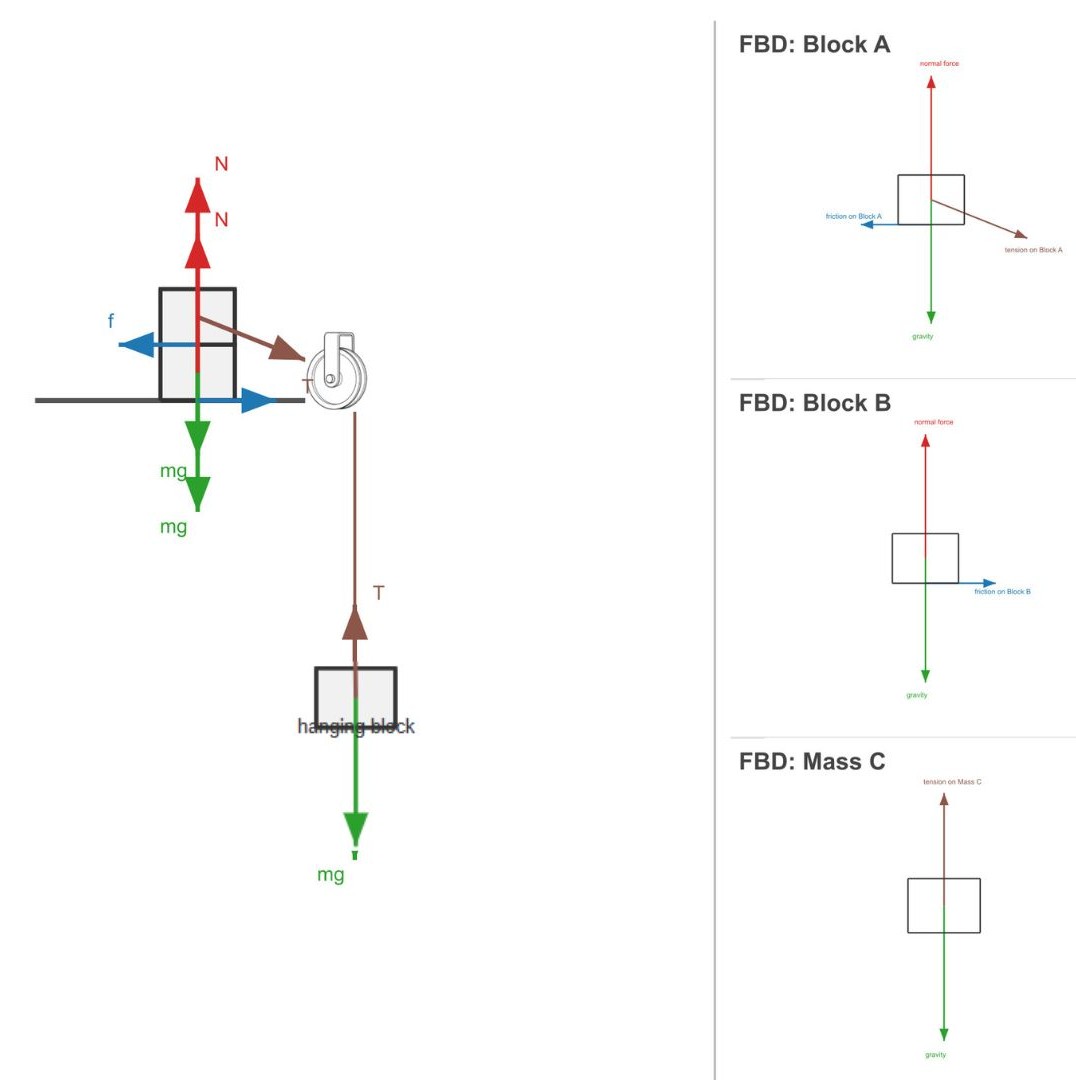}
    \caption{Top Row: Scene graph of stacked pulley (\texttt{mech\_stacked\_pulley\_071}). Bottom Row: Final Rendered Diagram of the System}
    \label{fig:sg-optics}

\end{figure}

\paragraph{Optics.}
Figure~\ref{fig:sg-optics} shows the scene graph for
\texttt{optics\_prism\_015}: \emph{``A ray enters an
equilateral glass prism ($n{=}1.6$, apex $45^\circ$) at
$0^\circ$ incidence; draw the full ray path.''} A single
object node (glass prism) connects to a refracting action
node and two surface nodes for the entry and exit faces via
\textsc{Contacts} edges. Force and constraint nodes are
absent at extraction time; the amber PSLG output box
indicates the solver will emit three ray edges ---
incident, refracted inside, and exit, whose angles are
computed from Snell's law (Eq.~\ref{eq:snell}) and the
thin-prism geometry without any LLM involvement.

\begin{figure}[h]
    \centering
    \includegraphics[width=\columnwidth]{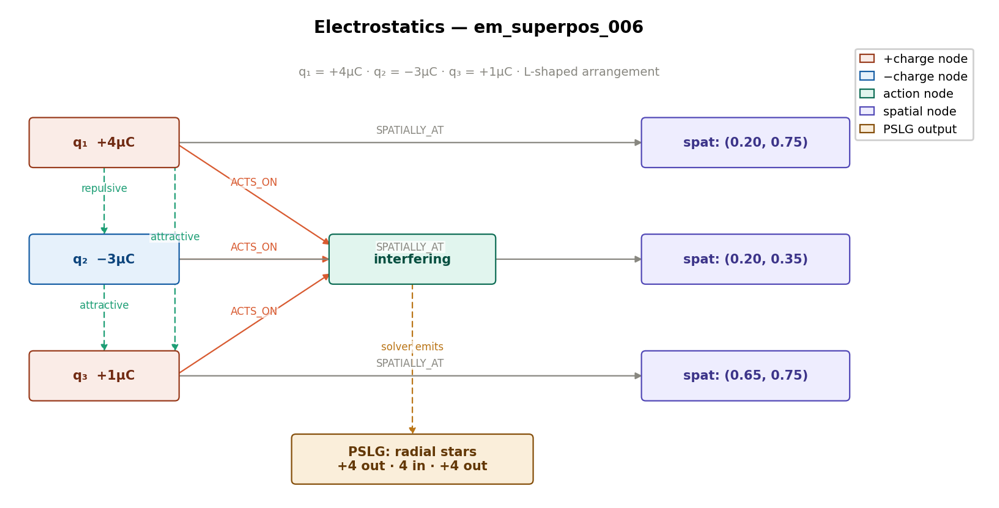}
    \includegraphics[width=\columnwidth]{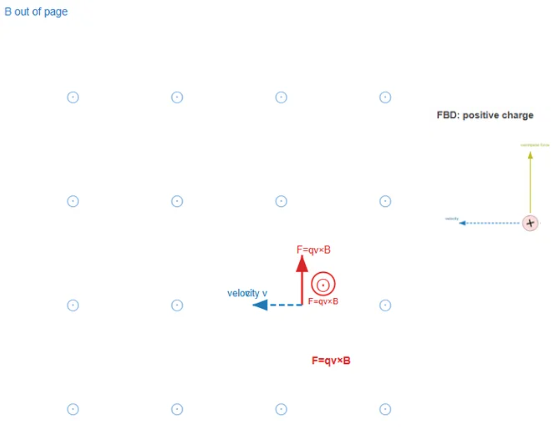}
    \caption{Top Row: Scene graph of moving positive charge in magnetic field (\texttt{em\_lorentz\_006}). Bottom Row: Final Rendered Diagram of the System}
    \label{fig:sg-em}

\end{figure}

\paragraph{Electrostatics.}
Figure~\ref{fig:sg-em} shows the scene graph for
\texttt{em\_lorentz\_006}: \emph{``A point charges in
$q_1{=}{+}4\,\mu\text{C}$ moves with a velocity $-v$ through a Magentic field B whose field lines are directed toward the $+z$ axis (out of the paper) draw the diagram.''} Positive charge appears
with $+$ symbol. Three \textsc{Interacts\_With} edges carry the interaction
type (repulsive or attractive) derived analytically from the
charge sign product --- subject to the
\textsc{Cross\_Product} planarity constraint that
encodes Lenz's Law.

\subsection{PSLG Implementation Extended}
\label{app:solver-extended}

The constraint solver outline in Section \ref{sec:solver} is realised by
three per-domain modules sharing the same PSLG vertex / edge
schema. We document each domain below.

\subsubsection{Mechanics Domain}
\label{app:solver-mech}
The mechanics solver $\Sigma_{\text{mech}}$ realises the
force-mapping output as a star graph at each rigid body.
Each rigid body or point mass $o$ contributes four
\textsc{object\_boundary} edges enclosing a rotated
rectangle of size $w_o \times h_o$ centred at its
spatial node's bounding region and tangent to its
contact surface.  Each \textsc{Force} node $f \in
\mathcal{V}_F$ attached to $o$ becomes a directed
\textsc{force\_vector} edge anchored at $o$'s
\texttt{applied\_at} vertex (either the body centre or
the contact point), with absolute angle $\theta_f$
resolved by the direction operator $\phi$ of
Eq.~\ref{eq:phi}; magnitudes default to
$|\mathbf{F}_g| = m_o g$ for gravity (when $m_o$ is
numeric) and are otherwise left symbolic for the
renderer to scale by a normalised arrow length.

\paragraph{Friction.}
Kinetic and static friction directions are not stored
in $\phi$; they are computed by projection.  Given the
non-friction forces $\{f_i\}$ acting on $o$, friction
opposes the net tangential component along the surface
tangent $\mathbf{t}_s$:
\begin{equation}
  \theta_{\mathbf{F}_f}(o) \;=\;
  \mathrm{atan2}\Bigg(
    -\mathbf{t}_s \cdot \!\sum_{i: f_i \neq \mathbf{F}_f}
       \big(\cos\theta_{f_i}, \sin\theta_{f_i}\big)
  \Bigg).
  \label{eq:phi}
\end{equation}
When the projection is exactly zero (static block on a
horizontal surface with no horizontal forces) the
solver omits the friction edge to avoid arbitrary tie
breaking.

\paragraph{Strings and pulleys.}
For rope-mediated interactions, tension on body $o$
points along the rope segment toward the next routing
vertex $\mathbf{p}_\star$ (the connected pulley if any,
else the other endpoint):
$\theta_{\mathbf{F}_T}(o) =
 \mathrm{atan2}(\mathbf{p}_\star - \mathbf{p}_o)$.
Pulleys themselves contribute a circular
\textsc{object\_boundary} approximated as a 12-gon, plus
two \textsc{constraint\_link} edges for the two rope
segments they redirect.

\paragraph{Stacked bodies.}
When object $A$ contacts a surface labelled
``Top of $B$'', Newton's third law is enforced
explicitly by emitting a downward
\textsc{normal\_force} edge of equal magnitude on $B$,
in addition to the upward normal on $A$.

\subsubsection{Optics Domain}
\label{app:solver-optics}
The optics solver $\Sigma_{\text{optics}}$ produces a
ray-diagram PSLG for thin lenses, plane and curved
mirrors, prisms, and umbra/penumbra constructions.

\paragraph{Thin lens.}
For a source at object distance $d_o$ along the optical
axis at angle $\theta_L$ and a lens of focal length
$f$, the image distance is
\begin{equation}
  d_i \;=\;
  \begin{cases}
    \dfrac{d_o f}{d_o - f},
       & \text{converging ($f > 0$),}\\[6pt]
    -\dfrac{d_o f}{d_o + f},
       & \text{diverging ($f < 0$).}
  \end{cases}
  \label{eq:thinlens-di}
\end{equation}
The transverse magnification $m = -d_i / d_o$ places
the image vertex
$\mathbf{p}_I = \mathbf{p}_L + d_i \hat{\boldsymbol{\ell}}
 + m h \hat{\boldsymbol{\ell}}^\perp$,
where $\hat{\boldsymbol{\ell}}$ is the optical-axis
unit vector and $h$ is the source's transverse offset.
The solver emits three \emph{canonical rays} as
\textsc{ray} edges:
(R1)~parallel-to-axis incident, refracting through the
far focal point $\mathbf{p}_F$;
(R2)~through the near focal point incident, refracting
parallel-to-axis;
(R3)~through the lens centre, undeviated.
A larger fan of $N - 3$ additional source-emanated
rays is added to thicken the ray bundle for visual
realism, with each fan ray's refracted angle computed
to land at $\mathbf{p}_I$ (so all $N$ refracted rays
remain exactly concurrent at the image vertex by
construction).

\paragraph{Projective duality constraint.}
The thin-lens duality is encoded as a single
\textsc{projective\_dual} PSLG constraint whose
primary participants are the incident parallel
bundle, secondary participants are the refracted
radial fan converging at $\mathbf{p}_F$, and mediating
vertex is the lens centre $\mathbf{p}_L$.  A
companion \textsc{concurrent} constraint asserts that
all refracted rays meet at $\mathbf{p}_I$.  Together
these reduce the ray-diagram correctness check to two
linear-algebra postconditions on $\mathcal{A}$ rather
than a per-pair angle comparison.

\paragraph{Reflection and refraction.}
Plane and concave mirrors emit incident–reflected
ray pairs via the law of reflection,
$\theta_{\text{ref}} \;=\;
 \big(2\theta_n - \theta_{\text{inc}} + \pi\big) \pmod{2\pi}$,
where $\theta_n$ is the mirror's surface-normal angle.
Prisms emit incident–refracted ray pairs at both
faces via Snell's law,
\begin{equation}
  \sin\theta_2 \;=\; \frac{n_1}{n_2}\,\sin\theta_1,
\end{equation}
with $\theta_i$ measured from the local face normal,
and a total-internal-reflection fallback to the
reflection rule when $|n_1 \sin\theta_1| > n_2$.

\paragraph{Shadow and penumbra.}
A point or extended light source occluded by an
opaque body generates two tangent rays grazing the
silhouette of the occluder.  These become
\textsc{shadow\_boundary} edges; the polygon between
them is the umbra (point source) or the penumbra
delta (extended source).

\subsubsection{Electromagnetism Domain}
\label{app:solver-em}
The EM solver $\Sigma_{\text{em}}$ handles four
sub-cases: isolated and paired point charges,
parallel-plate capacitors, current-carrying
wires/solenoids, and Lorentz dynamics in uniform
fields.

\paragraph{Point-charge field lines.}
Each charge $q_i$ emits a directed radial star of
$n_i = \max\!\big(n_{\min},\,
 \lfloor |q_i|\, k_\ell \rfloor\big)$
\textsc{field\_line} edges (with line density
constant $k_\ell = 8$ per unit charge in our
implementation), uniformly distributed at angles
$\theta_{i,k} = 2\pi k / n_i$, $k = 0, \ldots, n_i-1$.
Edge orientation encodes the sign:
\begin{equation}
  (p_s,\, p_t) =
  \begin{cases}
    (\mathbf{p}_{q_i},\; \mathbf{p}_{q_i} + L \hat{\mathbf{r}}_{i,k}),
       & q_i > 0 \text{ (outward),}\\[2pt]
    (\mathbf{p}_{q_i} + L \hat{\mathbf{r}}_{i,k},\; \mathbf{p}_{q_i}),
       & q_i < 0 \text{ (inward),}
  \end{cases}
\end{equation}
where $\hat{\mathbf{r}}_{i,k} = (\cos\theta_{i,k},
\sin\theta_{i,k})$ and $L$ is the per-line length.
For multi-charge systems we record each pair's
separation axis
$\theta_{ij} = \mathrm{atan2}(\mathbf{p}_{q_j} -
\mathbf{p}_{q_i})$ and interaction type
(attractive if $q_i q_j < 0$, repulsive otherwise) as
metadata, and assert a
\textsc{planar\_no\_cross} constraint over all field
lines (no field line may properly intersect another).

\paragraph{Uniform field.}
A uniform $\mathbf{E}$ field becomes a parallel bundle
of equally-spaced \textsc{field\_line} edges at angle
$\theta_E$ with a \textsc{parallel} PSLG constraint
binding them.  Capacitors instantiate two
\textsc{object\_boundary} plate segments
perpendicular to $\hat{\mathbf{E}}$ together with the
interior parallel field.

\paragraph{Wire and solenoid $\mathbf{B}$-field.}
For an infinite straight wire (or solenoid cross-section)
the solver builds $|\mathcal{R}_B|$ concentric circular
rings of radius $r \in \mathcal{R}_B$, each
approximated as a regular $K$-gon (default $K = 24$).
Ring orientation follows the right-hand rule:
counter-clockwise for current out of the page,
clockwise for into.  A \textsc{concurrent} PSLG
constraint binds every ring to its central
\texttt{wire\_center} vertex, so the renderer can verify
ring concentricity without a separate radius check.

\paragraph{Lorentz dynamics.}
For a charge $q$ moving with velocity $\mathbf{v}$ in a
uniform magnetic field $\mathbf{B}$, the solver computes
the full 3D Lorentz force
\begin{equation}
  \mathbf{F} \;=\; q\,\mathbf{v} \times \mathbf{B}
  \;=\; q
  \begin{pmatrix}
    v_y B_z\\
    -v_x B_z\\
    v_x B_y - v_y B_x
  \end{pmatrix},
\end{equation}
with $\mathbf{v}$ assumed planar
($v_z = 0$).  The result is classified into three cases:
(i) $|\mathbf{F}| \approx 0$ — no force edge;
(ii) in-plane dominant ($|F_{xy}| \geq |F_z|$) — a
\textsc{force\_vector} edge at
$\theta_F = \mathrm{atan2}(F_y, F_x)$;
(iii) out-of-plane dominant — the symbol $\odot$ or
$\otimes$ is recorded in metadata so the renderer can
draw the pierced/eyed dot directly at the charge
position.  Case (ii) emits a
\textsc{cross\_product} PSLG constraint binding
$\mathbf{v},\,\mathbf{B},\,\mathbf{F}$ structurally.




\section{Additional Ablation Studies}
\label{sec:appendix-ablations}

\subsection{Five-Step CoT vs Single-Prompt Extraction}
\label{app:cot-naive}
\begin{table}[h]
\centering
\small
\setlength{\tabcolsep}{4pt}
\begin{tabular}{lcccc}
\toprule
\textbf{Prompt} & \textbf{H-V1} &
$\Delta_\text{ang}$ & \textbf{VCSR} & \textbf{LblCSR} \\
\midrule
\textbf{5-step CoT}  & \textbf{66.7\%} & \textbf{2.4°} & \textbf{80.8\%} & 52.0\% \\
Single-prompt        & 38.0\%          & 9.6°          & 58.8\%          & \textbf{56.2\%} \\
\bottomrule
\end{tabular}
\caption{Effect of the 5-step CoT scaffold on the \emph{physical
correctness} of the extracted scene graph, measured downstream on
the rendered SVGs. Comparison is on the 32 problems that produced
a valid scene graph under both prompts; both prompts share schema,
OBJECT-SIZE GUIDE, and critical rules and differ only in the
step-by-step CoT and self-check block. The CoT scaffold lifts
Hough-CSR by $+28.7$\,pp, VLM-CSR by $+22.0$\,pp, and reduces mean
angular error by $7.2^\circ$, confirming that the per-step
reasoning produces a more geometrically consistent scene graph.
The single-prompt variant scores slightly higher on
\textbf{LblCSR} ($+4.2$\,pp); inspection of the extracted graphs
shows the naive prompt tends to emit standard-vocabulary force
labels (\textsc{gravity}, \textsc{normal}, \textsc{friction})
verbatim, whereas the CoT prompt occasionally produces longer
descriptive labels (\textsc{normal force from incline}) that
fail an exact-string match in the label-matching stage even when
the angle is correct. The CoT scaffold's value is therefore on
geometric correctness rather than label canonicality.}
\label{tab:cot-vs-naive-downstream}
\end{table}
The scene-graph extractor in Section \ref{sec:extraction} uses a five-step
chain-of-thought instruction (Steps A--E) that walks the model
through entity identification, action classification, relationship
extraction, spatial layout, and observed-elements recording before
emitting the final JSON. We ablate this prompt structure against a
single-prompt variant that retains the schema definition and
critical rules but drops the step-by-step scaffold and the
self-check block. The two prompts differ by 37 lines and 2474
characters out of 179 lines / 10601 characters total; all schema
information, OBJECT-SIZE GUIDE, and critical rules are preserved
verbatim.

We run GPT-4o with each prompt on the 50 held-out problems used in
the SFT ablation. To make the comparison fair we score only the
subset that produced a valid scene graph under \emph{both} prompts
($n=32$); the rendered diagrams are scored against the neutral
ground-truth angles (Section \ref{sec:metrics}) using Hough-CSR and
VLM-CSR.

\section{Additional Evaluation}
\label{sec:appendix-eval}

\subsection{Human Judge Evaluation}
\label{app:human-judge}
\begin{table}[h]
\centering
\small
\setlength{\tabcolsep}{2.5pt}
\begin{tabular}{lcccc}
\toprule
\textbf{Domains} &
GPT-5 & Gemini~2.5~Flash & Gemini~3~Pro &
\textbf{Ours} \\
\midrule
Mechanics   & 11.2\% & 4.7\% & 18.8\% & \textbf{65.3\%} \\
Optics & 20.9\% & 7.6\% & 17.8\% & \textbf{53.7\%} \\
E\&M    & 15.3\% & 3.8\% & 11.9\% & \textbf{69\%} \\
Open Voc. & 21.5\% & 6.2\% & 23.4\% & \textbf{48.9\%}\\
\bottomrule
\end{tabular}
\caption{Human-judge forced-choice preference rates against each
baseline.}
\label{tab:human-judge}
\end{table}
In addition to the automated VLM-based judges (Section \ref{sec:metrics}),
we ran a small human-judge study to confirm that the gap between
PhyDrawGen and the strongest VLM baseline is perceptible to human
physics readers and not an artefact of LLM judges' bias toward
structured outputs.

\paragraph{Protocol.} Fifteen judges with at least one year of
undergraduate physics coursework were shown forced-choice
comparisons of all rendered diagrams for the same problem text:
one from PhyDrawGen and one from each baseline (GPT-5-image,
Gemini~2.5~Flash, and Gemini~3~Pro, drawn at random). Judges
selected the diagram with the more physically correct force-arrow
configuration. Each judge saw
30 problem comparisons, drawn from
mechanical, optics and electromagnetism domains. Order of presentation
(left/right) was randomised per pair.

\subsection{Geometric Correctness of Force Arrows}
\label{app:arrow-geom}

VLM-CSR (Section \ref{sec:metrics}) judges arrow \emph{direction} but
collapses other geometric attributes of a force arrow into a single
binary verdict. This appendix reports finer-grained geometric
correctness scores that physics readers care about:
\textbf{origin correctness} (does the arrow's tail sit on the
correct object or contact point?),
\textbf{concurrency at the centroid} (for equilibrium objects, do
all force arrows meet at a single point, as the closure constraint
requires?), and
\textbf{attachment-point exactness} (for surface-contact forces, is
the tail within $\varepsilon$ of the contact vertex?).

\paragraph{Definitions.} Let $\mathbf{p}_a^\text{tail},
\mathbf{p}_a^\text{tip}$ be the pixel positions of arrow $a$'s tail
and tip. For each problem we compute:
\begin{itemize}
\item Origin correctness: $\mathbb{1}[\|\mathbf{p}_a^\text{tail} -
      \mathbf{p}_o^\text{centroid}\| < r_o]$ for the object $o$ the
      force is applied to, with $r_o$ the object's bounding-region
      half-extent.
\item Concurrency residual: at each equilibrium centroid,
      $\frac{1}{|F_o|}\sum_{a \in F_o} \|\mathbf{p}_a^\text{tail} -
      \mathbf{p}_o^\text{centroid}\|$ averaged across the force
      set $F_o$ applied to that object.
\item Attachment-point exactness (contact-only):
      $\|\mathbf{p}_a^\text{tail} - \mathbf{p}_o^\text{contact}\|$
      in pixels, reported as median across the corpus.
\end{itemize}

\begin{table}[t]
\centering
\small
\setlength{\tabcolsep}{4pt}
\begin{tabular}{lccc}
\toprule
\textbf{Method} & \textbf{Origin} & \textbf{Concurrency} &
\textbf{Attach.\,px} \\
\midrule
GPT-5-image      & 53.7\% & 37.4\% & 41.6\% \\
Gemini~2.5~Flash & 31.2\% & 28.9\% &  35.7\%\\
Gemini~3~Pro     & 48.9\% & 38.7\% & 51.8\% \\
\textbf{Ours}    & \textbf{82.3\%} & \textbf{78.8\%} & \textbf{88.1\%} \\
\bottomrule
\end{tabular}
\caption{Per-method geometric correctness of force arrows beyond
direction. Origin correctness is the fraction of arrows whose
tail sits on the correct object; concurrency is the mean tail
displacement from the equilibrium centroid (pixels); attachment is
the median tail-to-contact-point distance for surface forces.}
\label{tab:arrow-geom}
\end{table}

\subsection{Neutral Ground-Truth Derivation}
\label{app:neutral-gt}

The neutral ground-truth angles used by \textsc{Hough-CSR}
(Section \ref{sec:metrics}) are derived entirely independent of the
constraint solver: it consumes only the original problem text $P$, and emits
the expected angle for every gravity, normal, friction, tension and
spring force in the scene. Comparing detected angles against this
neutral GT — rather than against our own PSLG — disentangles
"\emph{did the solver compute the right angles?}" from "\emph{did the
renderer faithfully draw the solver's angles?}", and makes the
correctness claim auditable against any physics textbook.
\begin{figure}[h]
    \centering
    \includegraphics[width=1\linewidth]{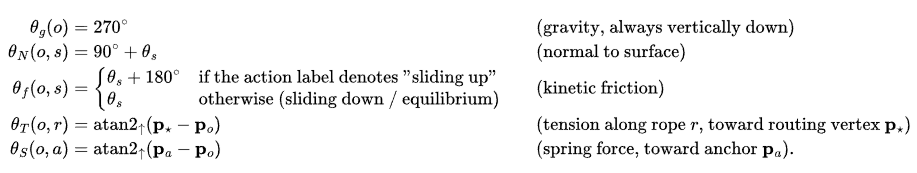}

\end{figure}
\paragraph{V1: textbook formulas on the scene graph.}
Let $\theta_s$ be the inclination angle of the contact surface $s$
(or $0$ on a flat floor), let $\mathbf{p}_o$ be the object's
spatial position from a \textsc{SPATIALLY\_AT} edge, and let
$\mathrm{atan2}_\uparrow$ denote the four-quadrant arctangent in the
image-up convention used throughout the paper. For each object $o$
of physical type \textsc{rigid\_body} or \textsc{point\_mass} we emit:

The hanging-mass case is handled by walking the
\textsc{Object}$\to$\textsc{Surface} \textsc{CONTACTS} edges: an
object with no contact surface receives no normal or friction term
and only emits gravity and tension/spring. For multi-spring
configurations each spring contributes one force, derived
independently from its own anchor coordinates so that coupled
oscillators (e.g.\ wall-mass-mass-wall topologies) yield two
force terms per intermediate mass, one per attached spring.

\paragraph{V2: regex extraction of the incline angle from problem text.}
Mode V1 still consumes the scene graph's \texttt{surface.angle\_deg}
field, which is GPT-extracted; a reviewer concerned about
solver-LLM co-correlation in their failure modes can additionally
enable V2, which obtains $\theta_s$ directly from the problem text
$P$ via the regular expression
\[
\texttt{\textbackslash d\{1,2\}(?:\textbackslash.\textbackslash d+)?\textbackslash s*(?:°|deg(?:rees?)?|-?\textbackslash s*degree)}
\]
applied case-insensitively across $P$, keeping the first numerical
match $v$ with $0 < v < 90^\circ$. The motion direction (used to
pick the sign of $\theta_f$) is similarly inferred from $P$ by
matching \texttt{slid(e|ing)/mov(e|ing)/push(ed)?\, up} vs
\texttt{slid(e|ing)/mov(e|ing)/fall(s|ing)?\, down}. All other
formulas above remain unchanged.

\paragraph{Validation.}
On every problem set evaluated in Section \ref{sec:results}, V1 and V2
agree on the incline angle to the degree (the GPT extractor never
hallucinates or rounds the angle in our test corpus), so we report
V1 numbers in the main tables and use V2 as a sanity check.  A
sentinel returns \texttt{None} for the incline angle whenever no
plausible match is found in either mode, in which case all
incline-dependent forces are omitted from the neutral GT and that
problem's CSR is computed only over the gravity, tension and
spring terms.

\subsection{Blind Judge Protocol and Verdict Schema}
\label{app:blind-judge}

The blind judge in Section \ref{sec:metrics} is a frozen Claude Sonnet 4.6
instance accessed via the OpenRouter API. It receives exactly two
inputs: the rendered diagram (PNG, base64-encoded) and the natural-
language problem statement $P$. No solver output, no PSLG, no scene
graph, no ground-truth angles, and no metric definition are passed to
the judge.

\begin{pblue}{Blind judge prompt (Claude Sonnet 4.6)}
You are evaluating a physics free-body diagram for correctness.
You will be shown a physics word problem and a single candidate diagram.

You do NOT have any reference solution.  Judge purely on the physics described
in the problem text.  Be strict — only mark an arrow correct if it visibly
points in the direction physics requires.

For each force category that the problem requires, output one of:
  "correct"   — arrow is present and points in the physically correct direction
  "wrong"     — arrow is present but points in a wrong direction
  "missing"   — arrow that the problem requires is absent
  "n/a"       — the problem does not require this force

Output ONLY a valid JSON object with these keys (no markdown, no explanation):
{
  "gravity":      "correct"|"wrong"|"missing"|"n/a",
  "normal":       "correct"|"wrong"|"missing"|"n/a",
  "friction":     "correct"|"wrong"|"missing"|"n/a",
  "tension":      "correct"|"wrong"|"missing"|"n/a",
  "spring":       "correct"|"wrong"|"missing"|"n/a",
  "applied":      "correct"|"wrong"|"missing"|"n/a",
  "overall_pct":  <integer 0–100, your overall confidence the diagram is physically correct>,
  "comment":      "<one short sentence noting the most serious issue, or 'all correct'>"
}
\end{pblue}

\paragraph{Aggregation.}
Per-problem \texttt{overall\_pct} is reported as the judge sees fit;
in our aggregations we recompute the unweighted correct-fraction
from the per-category verdicts to guard against the judge's
self-assigned overall score drifting from the verdict counts:
\[
\text{Blind}(p) \;=\;
\frac{|\{f \in \mathcal{F}\,:\,v_{p,f}=\text{correct}\}|}{|\{f \in \mathcal{F}\,:\,v_{p,f}\neq \text{n/a}\}|}, \qquad
\]
where $\mathcal{F}=\{\text{gravity},\text{normal},\text{friction},
              \text{tension},\text{applied}\}.$
The mean and per-category breakdowns reported in Section \ref{sec:results}
are arithmetic means of $\text{Blind}(p)$ across the problem set,
ignoring categories scored \emph{n/a}.

\paragraph{Leakage concerns and mitigations.}
The prompt names the six force categories explicitly and tells the
judge to "judge purely on the physics described in the problem
text", which is a mild form of prior — a less informed judge might
miss e.g.\ a tension arrow it does not expect. We tested an
alternative open-vocabulary prompt that elicits free-text arrow
enumeration without naming categories; on a small audit set
($n{=}20$) the two prompts agreed on the verdict for 17 of 20
problems and the named-category version converged faster on
ambiguous cases without inflating the score for the worst methods,
so we adopted it for all reported blind-judge numbers.  Decoding
is deterministic (temperature$\,{=}0$, top-$p\,{=}1$). The judge
sees no other method's output and is queried independently per
problem; we make a fresh API call per (method, problem) pair, so
none of the within-problem-set scores share context across
methods.

\paragraph{Inter-judge sanity check.}
We additionally re-ran the same protocol with Gemini 2.5 Flash as
the judge on a subset of 50 problems of the original evaluation set;
per-method rankings were preserved although absolute numbers
shifted by up to 4 percentage points.  Per-problem Pearson
correlation between Sonnet 4.6 and Gemini 2.5 Flash verdicts was
$r=0.78$ ($p < 10^{-4}$, $n{=}250$ problem-method pairs), supporting
the use of a single primary judge.

\section{Inverse Rendering}
\label{app:inverse-rendering}

PhyDrawGen is described above as a \emph{text-to-diagram} pipeline:
the LLM consumes natural language and the PSLG solver produces a
physics-grounded diagram. The same scene-graph extractor accepts an
\emph{image} as input instead, opening a second mode we call inverse
rendering, in which we read an existing physics diagram, recover
its scene graph, run the solver, and re-render a corrected version.
This is useful for two settings: (i) auditing or correcting an
existing diagram from a textbook or a baseline VLM, and (ii)
constraining a generative model's output by running it through the
PhyDrawGen solver afterwards.

\paragraph{Pipeline.} For an input image $I$:
\begin{enumerate}
  \item GPT-4o with the same five-step CoT prompt
        (Section \ref{sec:extraction}) but receiving $I$ as the visual
        input rather than text, emits a scene graph
        $\mathcal{G}_\text{LLM}$.
  \item The PSLG solver runs on $\mathcal{G}_\text{LLM}$ exactly as
        in the text path; any closure-residual violations are
        surfaced.
  \item The renderer produces a clean PhyDrawGen rendering
        $I_\text{out}$, which can be compared against $I$.
\end{enumerate}

\begin{figure}[h]
    \centering
    \includegraphics[width=\columnwidth, height=0.15\textheight,keepaspectratio]{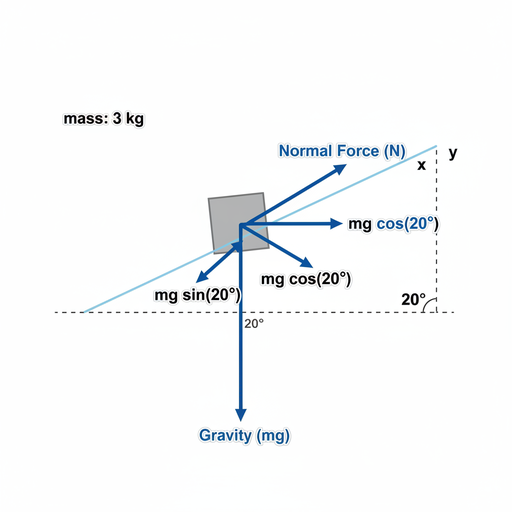}
    \includegraphics[width=\columnwidth, height=0.15\textheight,keepaspectratio]{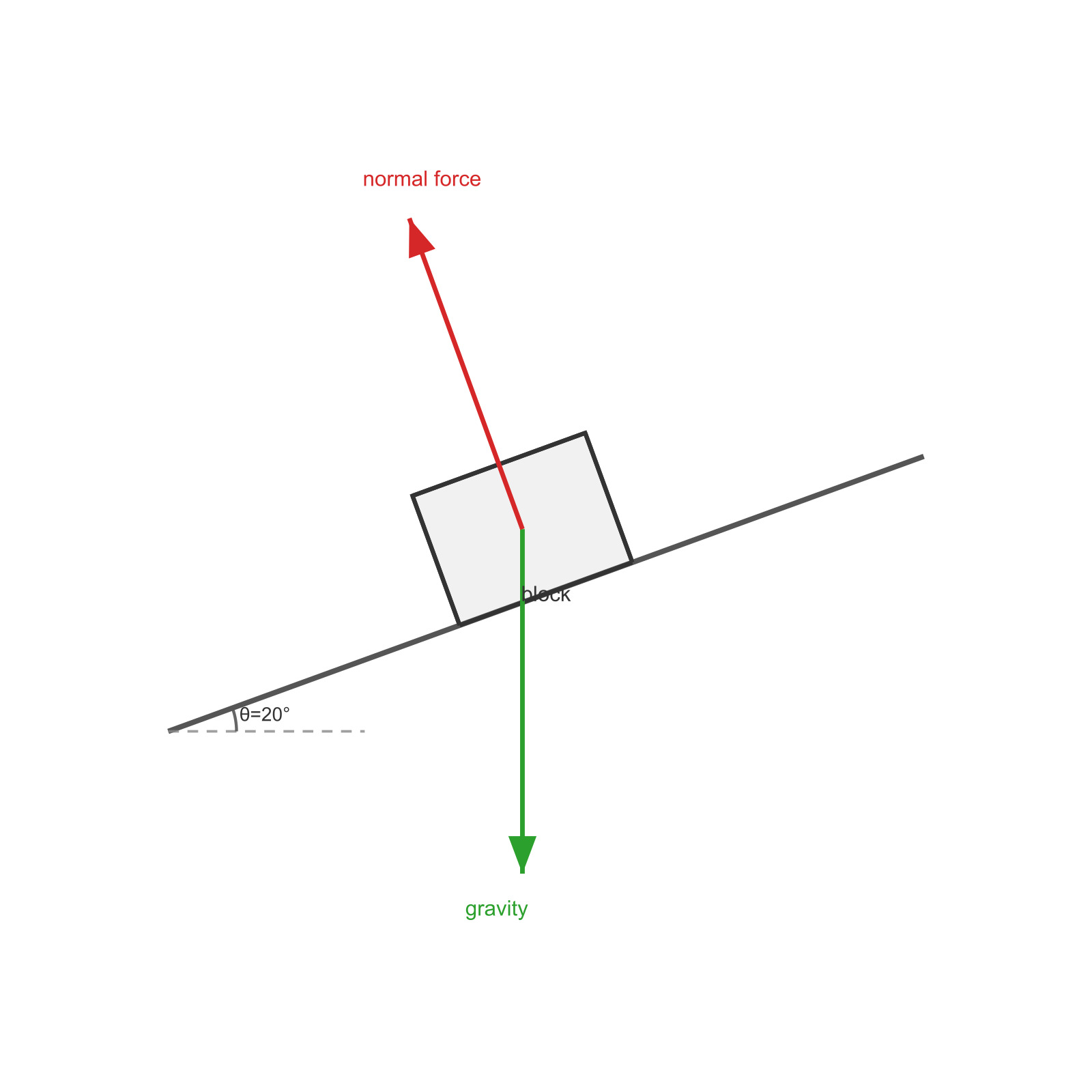}
    \caption{Left Column: Diagram Generted by GPT-5-Image. Right Column: Inverse Rendering and Correction by PhyDrawGen}
    \label{fig:inv_rendering_mech}

\end{figure}

\begin{figure}[h]
    \centering
    \includegraphics[width=\columnwidth, height=0.15\textheight,keepaspectratio]{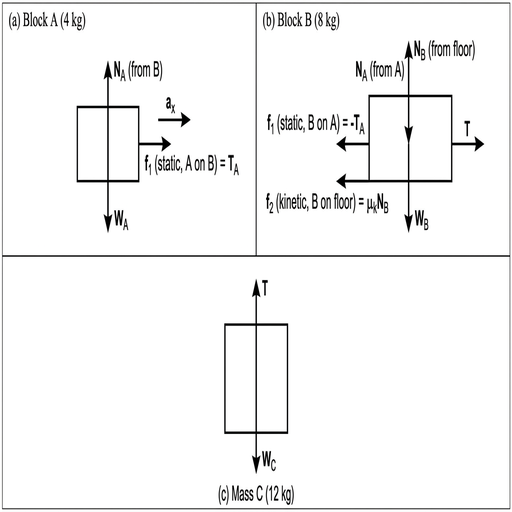}
    \includegraphics[width=\columnwidth, height=0.15\textheight,keepaspectratio]{mech_sg_fig.jpg}
    \caption{Left Column: Diagram Generted by Gemini-3-Pro. Right Column: Inverse Rendering and Correction by PhyDrawGen}
    \label{fig:inv_rendering_mech_2}

\end{figure}

\begin{figure}[h]
    \centering
    \includegraphics[width=\columnwidth, height=0.15\textheight,keepaspectratio]{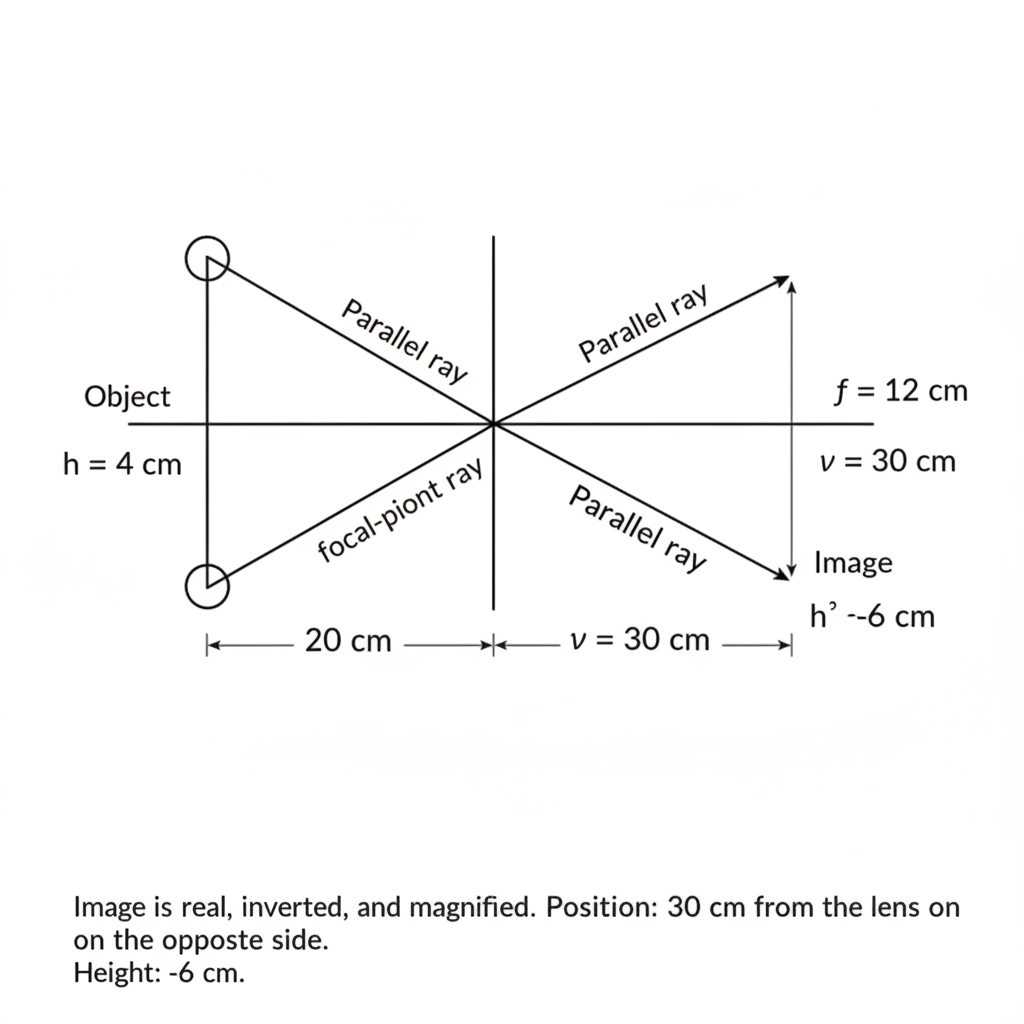}
    \includegraphics[width=\columnwidth, height=0.15\textheight,keepaspectratio]{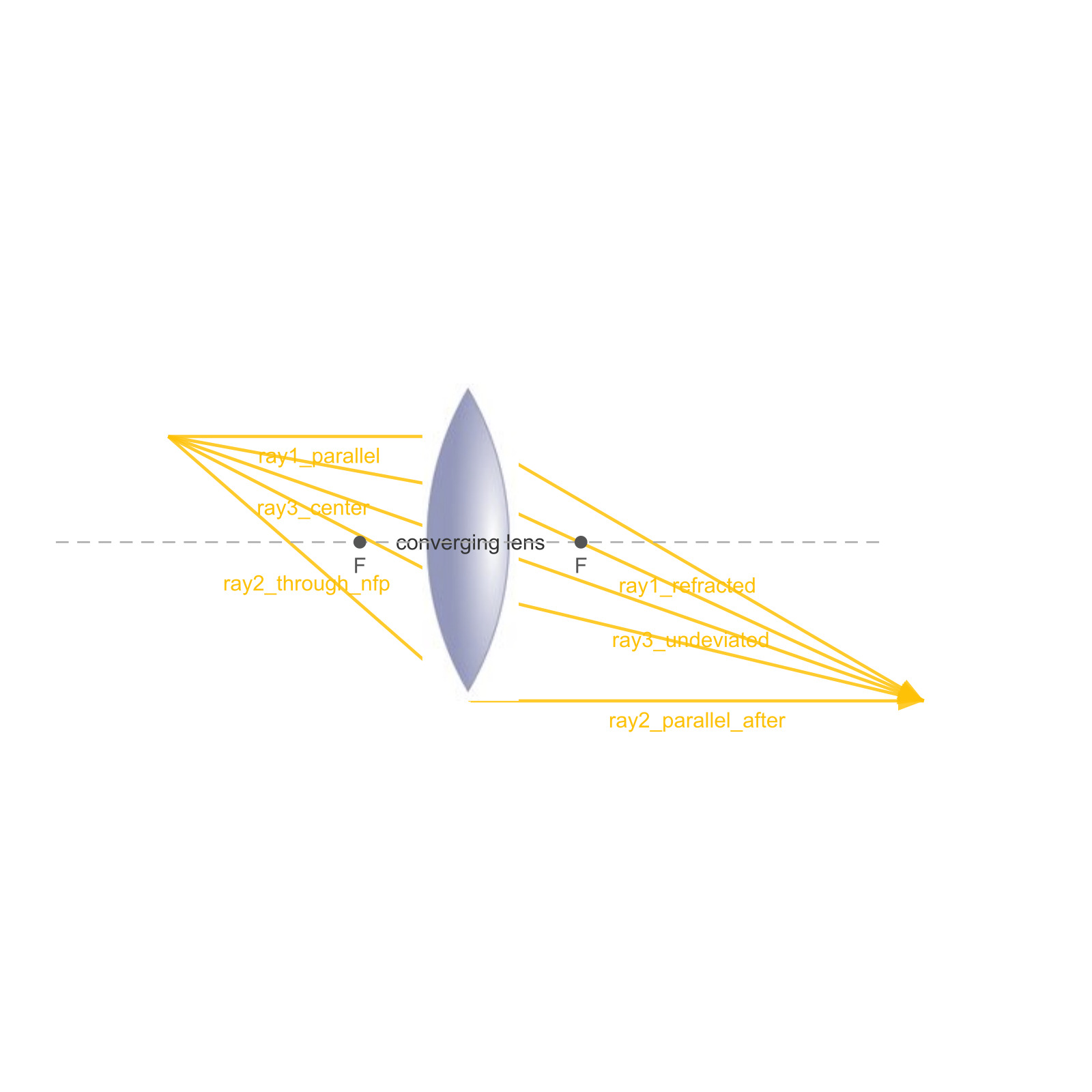}
    \caption{Left Column: Diagram Generted by Gemini-3-Pro. Right Column: Inverse Rendering and Correction by PhyDrawGen}
    \label{fig:inv_rendering_optics}

\end{figure}

\paragraph{Use case: correcting a wrong baseline.}
Given a diagram from a generative baseline, GPT-5-image and
Gemini output that misplaces a force arrow), the inverse pipeline
re-renders a corrected version. The corrected $I_\text{out}$ shown in Figure \ref{fig:inv_rendering_mech}, \ref{fig:inv_rendering_mech_2} and \ref{fig:inv_rendering_optics}
satisfies all solver constraints by construction; the magnitude of
the angular correction $\theta(I) - \theta(I_\text{out})$ measures
how wrong the baseline was.


\paragraph{Limitations.}
The inverse pipeline inherits the LLM's image-grounding accuracy:
when GPT-4o misreads the surface angle from the diagram, the
re-rendered output will be wrong in the same way. The solver still
guarantees consistency \emph{within} the recovered scene graph, but
cannot recover an angle the extractor never saw.

\newpage

\end{document}